\documentclass[11pt,a4paper]{article}
\usepackage[hyperref]{acl2018}
\usepackage{graphicx}
\usepackage{times}
\usepackage{latexsym}
\usepackage{url}
\usepackage{amsmath}
\usepackage{amssymb}
\usepackage{float}
\usepackage{multirow}
\usepackage{rotating}

\def \hfillx {\hspace*{-\textwidth} \hfill}

\aclfinalcopy 
\def\aclpaperid{625} 

\title{Mem2Seq: Effectively Incorporating Knowledge Bases into End-to-End Task-Oriented Dialog Systems}

\author{Andrea Madotto\thanks{$^*$ These two authors contributed equally.} , Chien-Sheng Wu$^*$, Pascale Fung\\
  Human Language Technology Center \\
  Center for Artificial Intelligence Research (CAiRE) \\
  Department of Electronic and Computer Engineering \\
  The Hong Kong University of Science and Technology, Clear Water Bay, Hong Kong \\
  {\tt [eeandreamad,cwuak,pascale]@ust.hk} }

\date{}

\begin{document}
\maketitle
\begin{abstract}
End-to-end task-oriented dialog systems usually suffer from the challenge of incorporating knowledge bases. 
In this paper, we propose a novel yet simple end-to-end differentiable model called memory-to-sequence (Mem2Seq) to address this issue. Mem2Seq is the first neural generative model that combines the multi-hop attention over memories with the idea of pointer network. We empirically show how Mem2Seq controls each generation step, and how its multi-hop attention mechanism helps in learning correlations between memories. In addition, our model is quite general without complicated task-specific designs. As a result, we show that Mem2Seq can be trained faster and attain the state-of-the-art performance on three different task-oriented dialog datasets.
\end{abstract}

\begin{figure*}[t]
\centering
\includegraphics[width=\linewidth]{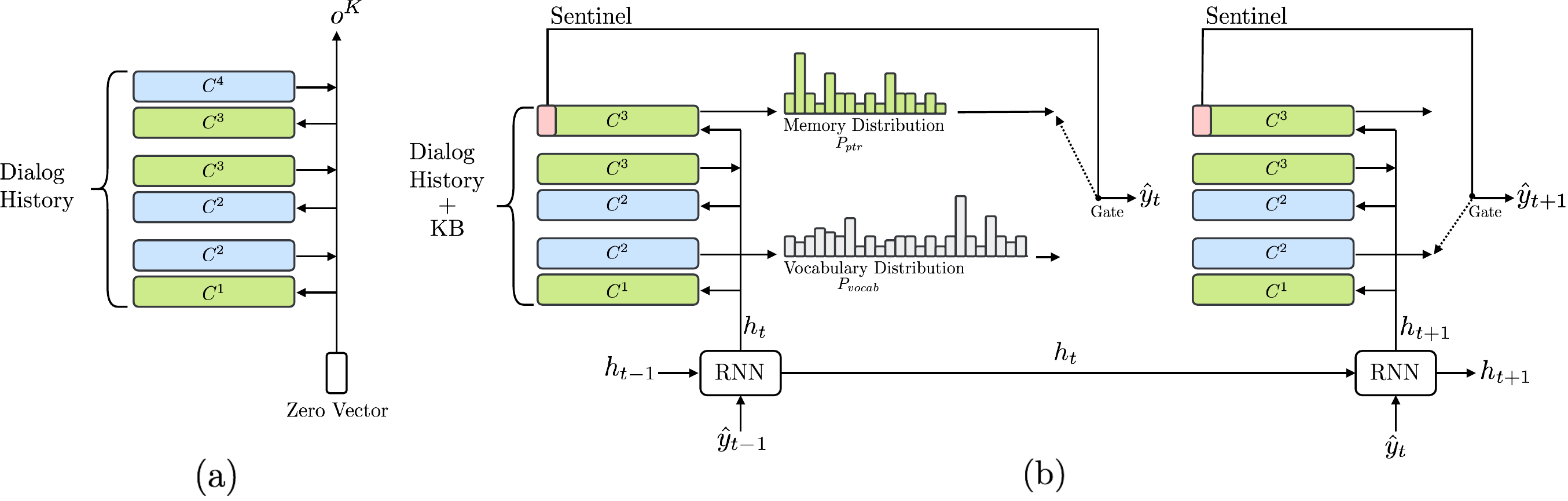}
\setlength{\abovecaptionskip}{-10pt} 
\caption{The proposed Mem2Seq architecture for task-oriented dialog systems. (a) Memory encoder with 3 hops; (b) Memory decoder over 2 step generation.}
\label{FIG:MODEL}
\end{figure*}

\section{Introduction}

\begin{table}[t]
\centering
\resizebox{\linewidth}{!}{
\begin{tabular}{|c|c|c|c|c|}
\hline
\textbf{Point of interest (poi)} & \textbf{Distance} & \textbf{Traffic info} & \textbf{Poi type} & \textbf{Address} \\ \hline
 The Westin & 5 miles           & moderate traffic       & rest stop           & 329 El Camino Real              \\ \hline
Round Table  & 4 miles           & no traffic             & pizza restaurant    & 113 Anton Ct                 \\ \hline
Mandarin Roots & 5 miles           & no traffic             & chinese restaurant  & 271 Springer Street        \\ \hline
Palo Alto Cafe & 4 miles           & moderate traffic       & coffee or tea place & 436 Alger Dr                \\ \hline
Dominos & 6 miles           & heavy traffic          & pizza restaurant    & 776 Arastradero Rd                 \\ \hline
Stanford Express Care & 6 miles           & no traffic             & hospital            & 214 El Camino Real   \\ \hline
Hotel Keen & 2 miles           & heavy traffic          & rest stop           & 578 Arbol Dr                    \\ \hline
\end{tabular}
}
\resizebox{\linewidth}{!}{
\begin{tabular}{l|rl}
\hline
\multirow{6}{*}{\begin{sideways}\textit{1th Turn}\end{sideways}} & \textbf{DRIVER} & Where can I get tea? \\ \cline{2-3} 
 & \textit{Seq2Seq} & I have a away from away would you like the address \\ 
 & \textit{+Attn} & I have a listing for a place that serves tea that is 5 miles away \\ 
 & \textit{Ptr-Unk} & There is a away you like would you like more info \\ \cline{2-3} 
 & \textit{\textbf{Mem2Seq}} & \begin{tabular}[c]{@{}l@{}}The nearest tea shop is Palo Alto Cafe located 4 miles away, \\ would you like directions there?\end{tabular} \\ \cline{2-3} 
 & \textbf{GOLD} & \begin{tabular}[c]{@{}l@{}}Palo Alto Cafe is 4 miles away and serves coffee and tea. \\ Do you want the address?\end{tabular} \\ \hline
\multirow{6}{*}{\begin{sideways}\textit{2th Turn}\end{sideways} } & \textbf{DRIVER} & Yes. \\ \cline{2-3} 
 & \textit{Seq2Seq} & The is at at \\ 
 & \textit{+Attn} & \begin{tabular}[c]{@{}l@{}}The address is 329 El Camino Real and it's 3 miles away \\ and there is no traffic\end{tabular} \\ 
 & \textit{Ptr-Unk} & Palo Alto Cafe is 4 miles away and PAD is no traffic \\ \cline{2-3} 
 & \textit{\textbf{Mem2Seq}} & Palo Alto Cafe is 4 miles away at 436 Alger Drive \\ \cline{2-3} 
 & \textbf{GOLD} & Palo Alto is located at 436 Alger Dr. \\ \hline
\end{tabular}
}
\setlength{\abovecaptionskip}{-3pt} 
\caption{Example of generated responses for the In-Car Assistant on the navigation domain.}
\label{EXAMPL_DIALOG_1}
\end{table}

Task-oriented dialog systems help users to achieve specific goals with natural language such as restaurant reservation and schedule arrangement. Traditionally, they have been built with several pipelined modules: language understanding, dialog management, knowledge query, and language generation~\cite{williams2007partially,hori2009statistical,lee2009example, levin2000stochastic,young2013pomdp}. Moreover, the ability to query external Knowledge Bases (KBs) is essential in task-oriented dialog systems, since the responses are guided not only by the dialog history but also by the query results (e.g. Table~\ref{EXAMPL_DIALOG_1}). However, despite the stability of such pipelined systems via combining domain-specific knowledge and slot-filling techniques, modeling the dependencies between modules is complex and the KB interpretation requires human effort.

Recently, end-to-end approaches for dialog modeling, which use recurrent neural networks (RNN) encoder-decoder models, have shown promising results~\cite{serban2016building,wen2016network,zhao2017generative}. Since they can directly map plain text dialog history to the output responses, and the dialog states are latent, there is no need for hand-crafted state labels. Moreover, attention-based copy mechanism~\cite{gulcehre-EtAl:2016:P16-1,eric-manning:2017:EACLshort} have been recently introduced to copy words directly from the input sources to the output responses. Using such mechanism, even when unknown tokens appear in the dialog history, the models are still able to produce correct and relevant entities.

However, although the above mentioned approaches were successful, they still suffer from two main problems: 
1) They struggle to effectively incorporate external KB information into the RNN hidden states~\cite{sukhbaatar2015end}, since RNNs are known to be unstable over long sequences.
2) Processing long sequences is very time-consuming, especially when using attention mechanisms.

On the other hand, end-to-end memory networks (MemNNs) are recurrent attention models over a possibly large external memory~\cite{sukhbaatar2015end}. They write external memories into several embedding matrices, and use query vectors to read memories repeatedly. This approach can memorize external KB information and rapidly encode long dialog history. Moreover, the multi-hop mechanism of MemNN has empirically shown to be essential in achieving high performance on reasoning tasks~\cite{bordes2016learning}. Nevertheless, MemNN simply chooses its responses from a predefined candidate pool rather than generating word-by-word. In addition, the memory queries need explicit design rather than being learned, and the copy mechanism is absent.

To address these problems, we present a novel architecture that we call Memory-to-Sequence (Mem2Seq) to learn task-oriented dialogs in an end-to-end manner. In short, our model augments the existing MemNN framework with a sequential generative architecture, using global multi-hop attention mechanisms to copy words directly from dialog history or KBs. We summarize our main contributions as such: 
1) Mem2Seq is the first model to combine multi-hop attention mechanisms with the idea of pointer networks, which allows us to effectively incorporate KB information. 
2) Mem2Seq learns how to generate dynamic queries to control the memory access. In addition, we visualize and interpret the model dynamics among hops for both the memory controller and the attention.
3) Mem2Seq can be trained faster and achieve state-of-the-art results in several task-oriented dialog datasets.

\section{Model Description}
Mem2Seq~\footnote{The code is available at \url{https://github.com/HLTCHKUST/Mem2Seq} } is composed of two components: the MemNN encoder, and the memory decoder as shown in Figure~\ref{FIG:MODEL}.
The MemNN encoder creates a vector representation of the dialog history. Then the memory decoder reads and copies the memory to generate a response. 
We define all the words in the dialog history as a sequence of tokens $X=\{ x_1,\dots,x_n,\$\}$, where $\$$ is a special charter used as a sentinel, and the KB tuples as $B=\{ b_1,\dots,b_{l} \}$. We further define $U=[B; X]$ as the concatenation of the two sets $X$ and $B$, $Y=\{ y_1,\dots,y_m\}$ as the set of  words in the expected system response, and $PTR=\{ {ptr}_1,\dots,{ptr}_m\}$ as the pointer index set:
\begin{equation}
{ptr}_i = 
\begin{cases} 
max(z) &\mbox{if } \exists z \ \text{s.t.} \ y_i = u_z \\ 
n+l+1&\mbox{otherwise} 
\end{cases} 
\label{eq1}
\end{equation}
where $u_z\in U$ is the input sequence and $n+l+1$ is the sentinel position index. 

\subsection{Memory Encoder}
Mem2Seq uses a standard MemNN with adjacent weighted tying~\cite{sukhbaatar2015end} as an encoder. The input of the encoder is word-level information in $U$. The memories of MemNN are represented by a set of trainable embedding matrices $C = \{ C^1,\dots,C^{K+1} \}$, where each $C^k$ maps tokens to vectors, and a query vector $q^k$ is used as a reading head. The model loops over $K$ hops and it computes the attention weights at hop $k$ for each memory $i$ using:
\begin{equation}
  p^k_i = \text{Softmax}((q^k)^T C^k_i),
  \label{eq2}
\end{equation}
where $C^k_i = C^k(x_i)$ is the memory content in position $i$, and $\text{Softmax}(z_i)=e^{z_i}/\Sigma_j {e^{z_j}}$. Here, $p^k$ is a soft memory selector that decides the memory relevance with respect to the query vector $q^k$. Then, the model reads out the memory $o^k$ by the weighted sum over $C^{k+1}$~\footnote{Here is $C^{k+1}$ since we use adjacent weighted tying.},
\begin{equation}
  o^k = \sum_i p^k_i C^{k+1}_i.
  \label{eq3}
\end{equation}
Then, the query vector is updated for the next hop by using $q^{k+1} = q^{k} + o^{k}$. The result from the encoding step is the memory vector $o^K$, which will become the input for the decoding step.  

\subsection{Memory Decoder}
The decoder uses RNN and MemNN. The MemNN is loaded with both $X$ and $B$, since we use both dialog history and KB information to generate a proper system response. A Gated Recurrent Unit (GRU)~\cite{GRU}, is used as a dynamic query generator for the MemNN. At each decoding step $t$, the GRU gets the previously generated word and the previous query as input, and it generates the new query vector. Formally:
\begin{equation}
h_t = \text{GRU}(C^1(\hat{y}_{t-1}), h_{t-1});
\label{eq4}
\end{equation}
Then the query $h_t$ is passed to the MemNN which will produce the token, where $h_0$ is the encoder vector $o^K$. At each time step, two distribution are generated: one over all the words in the vocabulary ($P_{vocab}$), and one over the memory contents ($P_{ptr}$), which are the dialog history and KB inofrmation. 
The first, $P_{vocab}$, is generated by concatenating the first hop attention read out and the current query vector. 
\begin{equation}
P_{vocab}(\hat{y_t}) = \text{Softmax}(W_1[h_t;o^1]) 
\label{eq5}
\end{equation}
where $W_1$ is a trainable parameter. On the other hand, $P_{ptr}$ is generated using the attention weights at the last MemNN hop of the decoder: $P_{ptr} = p_t^K$. Our decoder generates tokens by pointing to the input words in the memory, which is a similar mechanism to the attention used in pointer networks~\cite{NIPS2015_5866}.

We designed our architecture in this way because we expect the attention weights in the first and the last hop to show a ``looser'' and ``sharper'' distribution, respectively. To elaborate, the first hop focuses more on retrieving memory information and the last one tends to choose the exact token leveraging the pointer supervision. Hence, during training all the parameters are jointly learned by minimizing the sum of two standard cross-entropy losses: one between $P_{vocab}(\hat{y_t})$ and $y_t\in Y$ for the vocabulary distribution, and one between $P_{ptr}(\hat{y_t})$ and ${ptr}_t\in PTR$ for the memory distribution. 

\subsubsection{Sentinel}
If the expected word is not appearing in the memories, then the $P_{ptr}$ is trained to produce the sentinel token $\$$, as shown in Equation~\ref{eq1}. Once the sentinel is chosen, our model generates the token from $P_{vocab}$, otherwise, it takes the memory content using the $P_{ptr}$ distribution. Basically, the sentinel token is used as a hard gate to control which distribution to use at each time step. A similar approach has been used in~\cite{merity2016pointer} to control a soft gate in a language modeling task. With this method, the model does not need to learn a gating function separately as in~\citet{gulcehre-EtAl:2016:P16-1}, and is not constrained by a soft gate function as in~\citet{see-liu-manning:2017:Long}. 

\subsection{Memory Content}
We store word-level content $X$ in the memory module. Similar to ~\citet{bordes2016learning}, we add temporal information and speaker information in each token of $X$ to capture the sequential dependencies. For example, ``hello $t1$ $\$u$'' means ``hello'' at time step 1 spoken by a user.

On the other hand, to store $B$, the KB information, we follow the works of~\citet{miller-EtAl:2016:EMNLP2016,ericKVR2017} that use a (subject, relation, object) representation. 
For example, we represent the information of The Westin in Table~\ref{EXAMPL_DIALOG_1}: (The Westin, Distance, 5 miles). Thus, we sum word embeddings of the subject, relation, and object to obtain each KB memory representation. During decoding stage, the object part is used as the generated word for $P_{ptr}$. For instance, when the KB tuple (The Westin, Distance, 5 miles) is pointed, our model copies ``5 miles'' as an output word. Notice that only a specific section of the KB, relevant to a specific dialog, is loaded into the memory.

\begin{table}[t]
\centering
\resizebox{\linewidth}{!}{
\begin{tabular}{r|ccccc|c|c}
\hline
\textbf{Task} & \textbf{1} & \textbf{2} & \textbf{3} & \textbf{4} & \textbf{5} & \textbf{DSTC2} & \textbf{In-Car} \\ \hline
\textit{Avg. User turns} & 4 & 6.5 & 6.4 & 3.5 & 12.9 & 6.7 & 2.6 \\
\textit{Avg. Sys turns} & 6 & 9.5 & 9.9 & 3.5 & 18.4 & 9.3 & 2.6 \\
\textit{Avg. KB results} & 0 & 0 & 24 & 7 & 23.7 & 39.5 & 66.1 \\ 
\textit{Avg. Sys words} & 6.3 & 6.2 & 7.2 & 5.7 & 6.5 & 10.2 & 8.6 \\ 
\textit{Max. Sys words} & 9 & 9 & 9 & 8 & 9 & 29 & 87 \\
\textit{Pointer Ratio} & .23 & .53 & .46 & .19 & .60 & .46 & .42\\ \hline
\textit{Vocabulary} & \multicolumn{5}{c|}{3747} & 1229 & 1601 \\
\textit{Train dialogs} & \multicolumn{5}{c|}{1000} & 1618 & 2425 \\
\textit{Val dialogs} & \multicolumn{5}{c|}{1000}  & 500 & 302 \\
\textit{Test dialogs} & \multicolumn{5}{c|}{1000 + 1000 OOV} & 1117 & 304 \\ \hline
\end{tabular}
}
\setlength{\abovecaptionskip}{-3pt} 
\caption{Dataset statistics for 3 different datasets.}
\label{Tab:statistic}
\end{table}

\begin{table*}[t]
\centering
\resizebox{\linewidth}{!}{
\begin{tabular}{r|ccc|ccc|ccc}
\hline
\textbf{Task} & QRN & MemNN & GMemNN & Seq2Seq & Seq2Seq+Attn & Ptr-Unk & \textbf{\begin{tabular}[c]{@{}c@{}}Mem2Seq H1\end{tabular}} & \textbf{\begin{tabular}[c]{@{}c@{}}Mem2Seq H3\end{tabular}} & \textbf{\begin{tabular}[c]{@{}c@{}}Mem2Seq H6\end{tabular}} \\ \hline
\textit{T1} 	& 99.4 (-) 	& 99.9 (99.6) 	& 100 (100) & 100 (100) & 100 (100) & 100 (100) & 100 (100) & 100 (100) & 100 (100) \\
\textit{T2} 	& 99.5 (-)	& 100 (100) 	& 100 (100) & 100 (100) & 100 (100) & 100 (100) & 100 (100) & 100 (100) & 100 (100) \\
\textit{T3} 	& 74.8 (-)	& 74.9 (2.0) 	& 74.9 (0) 	& 74.8 (0) & 74.8 (0) & 85.1 (19.0) & 87.0 (25.2)    & 94.5 (59.6)    & \textbf{94.7} (\textbf{62.1}) \\
\textit{T4} 	& 57.2 (-) 	& 59.5 (3.0) 	& 57.2 (0)  & 57.2 (0) & 57.2 (0) & {100} (100) & 97.6 (91.7) & {100} (100)  & \textbf{100} (\textbf{100}) \\
\textit{T5} 	& \textbf{99.6} (-) & 96.1 (49.4) 		& 96.3 (52.5) & 98.8 (81.5) & 98.4 (87.3) & 99.4 (91.5) & 96.1 (45.3) & 98.2 (72.9) & 97.9 (69.6) \\ \hline
\textit{T1-OOV} & 83.1 (-) 	& 72.3 (0) 		& 82.4 (0) 	& 79.9 (0) &    81.7 (0) &    92.5 (54.7) &    93.4 (60.4) &    91.3 (52.0) &    \textbf{94.0} (\textbf{62.2}) \\
\textit{T2-OOV} & 78.9 (-)  & 78.9 (0) 		& 78.9 (0) 	& 78.9 (0) &    78.9 (0) &    83.2 (0) &    81.7 (1.2) &    84.7 (7.3) &    \textbf{86.5} (\textbf{12.4}) \\
\textit{T3-OOV} & 75.2 (-) 	& 74.4 (0) 		& 75.3 (0) 	& 74.3 (0) &    75.3 (0) &    82.9 (13.4) &    86.6 (26.2) &    \textbf{93.2} (\textbf{53.3}) &    90.3 (38.7) \\
\textit{T4-OOV} & 56.9 (-) 	& 57.6 (0) 		& 57.0 (0) 	& 57.0 (0) &    57.0 (0) &    100 (100)  &    97.3 (90.6) &    100 (100) &    \textbf{100} (\textbf{100})\\
\textit{T5-OOV} & 67.8 (-) 	& 65.5 (0) 		& 66.7 (0) 	& 67.4 (0) &    65.7 (0) &    73.6 (0) &    67.6 (0) &  78.1 (0.4)  &    \textbf{84.5} (\textbf{2.3}) \\ \hline
\end{tabular}}
\setlength{\abovecaptionskip}{-3pt} 
\caption{Per-response and per-dialog (in the parentheses) accuracy on bAbI dialogs. Mem2Seq achieves the highest average per-response accuracy and has the least out-of-vocabulary performance drop. }
\label{TB:babi-dialog}
\end{table*}

\begin{table*}[t]
\begin{minipage}{0.39\textwidth}
\centering
\resizebox{\linewidth}{!}{
\begin{tabular}{r|c|c|c|c}
\hline
\multicolumn{1}{l|}{\textbf{}} & \textbf{Ent. F1} & \textbf{BLEU} & \textbf{\begin{tabular}[c]{@{}c@{}}Per-\\ Resp.\end{tabular}} & \textbf{\begin{tabular}[c]{@{}c@{}}Per-\\ Dial.\end{tabular}} \\ \hline
\textit{Rule-Based}            & -             & -              & 33.3         & -             \\
\textit{QRN}                   & -             & -              & 43.8         & -             \\
\textit{MemNN}                 & -             & -              & 41.1         & 0.0           \\
\textit{GMemNN}                & -             & -              & \textbf{47.4}& 1.4           \\
\textit{Seq2Seq}               & 69.7          & 55.0           & 46.4         & \textbf{1.5}  \\
\textit{+Attn}                 & 67.1          & \textbf{56.6}  & 46.0         & 1.4           \\
\textit{+Copy}                 & 71.6          & 55.4           & 47.3         & 1.3           \\ \hline
\textit{\textbf{Mem2Seq H1}}   & 72.9          & 53.7           & 41.7         & 0.0           \\
\textit{\textbf{Mem2Seq H3}}   & \textbf{75.3} & 55.3          & 45.0	       & 0.5          \\
\textit{\textbf{Mem2Seq H6}}   & 72.8          & 53.6          & 42.8	       & 0.7           \\ \hline
\end{tabular}}
\setlength{\abovecaptionskip}{-3pt} 
\caption{Evaluation on DSTC2. Seq2Seq (+attn and +copy) is reported from ~\citet{eric-manning:2017:EACLshort}.}
\label{TB:DSTC2}
\end{minipage} 
\hfillx
\begin{minipage}{0.58\textwidth}
\centering
\resizebox{\linewidth}{!}{
\begin{tabular}{r|c|c|ccc}
\hline
\multicolumn{1}{l|}{\textbf{}} & \textbf{BLEU}  & \textbf{Ent. F1} & \textbf{Sch. F1} & \textbf{Wea. F1} & \textbf{Nav. F1} \\ \hline
\textit{Human*} & 13.5 & 60.7 & 64.3 & 61.6 & 55.2  \\
\textit{Rule-Based*} & 6.6 & 43.8 & 61.3 & 39.5 & 40.4 \\ 
\textit{KV Retrieval Net*} & 13.2 & 48.0 & 62.9 & 47.0 & 41.3 \\ \hline\hline
\textit{Seq2Seq} & 8.4 & 10.3 & 09.7 & 14.1 & 07.0  \\
\textit{+Attn} & 9.3 & 19.9 & 23.4 & 25.6 & 10.8  \\
\textit{Ptr-Unk}  & 8.3 & 22.7 & 26.9 & 26.7 & 14.9  \\ \hline
\textit{\textbf{Mem2Seq H1}} & 11.6 & 32.4 & 39.8 & \textbf{33.6} & \textbf{24.6} \\
\textit{\textbf{Mem2Seq H3}} & \textbf{12.6} & \textbf{33.4} & \textbf{49.3} & 32.8 & 20.0  \\
\textit{\textbf{Mem2Seq H6}} & 9.9 & 23.6 & 34.3 & 33.0 & 4.4 \\
\hline
\end{tabular}}
\setlength{\abovecaptionskip}{-3pt} 
\caption{Evaluation on In-Car Assistant. Human, rule-based and KV Retrieval Net evaluation (with *) are reported from~\cite{ericKVR2017}, which are not directly comparable. Mem2Seq achieves highest BLEU and entity F1 score over baselines.}
\label{TB:INCAR}
\end{minipage}
\end{table*}   
\section{Experimental Setup}

\subsection{Dataset}
We use three public multi-turn task-oriented dialog datasets to evaluate our model: the bAbI dialog~\cite{bordes2016learning}, DSTC2~\cite{henderson2014second} and In-Car Assistant~\cite{ericKVR2017}. The train/validation/test sets of these three datasets are split in advance by the providers. The dataset statistics are reported in Table~\ref{Tab:statistic}. 

The bAbI dialog includes five end-to-end dialog learning tasks in the restaurant domain, which are simulated dialog data. Task 1 to 4 are about API calls, refining API calls, recommending options, and providing additional information, respectively. Task 5 is the union of tasks 1-4. There are two test sets for each task: one follows the same distribution as the training set and the other has out-of-vocabulary (OOV) entity values that does not exist in the training set.

We also used dialogs extracted from the Dialog State Tracking Challenge 2 (DSTC2) with the refined version from ~\citet{bordes2016learning}, which ignores the dialog state annotations. The main difference with bAbI dialog is that this dataset is extracted from real human-bot dialogs, which is noisier and harder since the bots made mistakes due to speech recognition errors or misinterpretations.

Recently, In-Car Assistant dataset has been released. which is a human-human, multi-domain dialog dataset collected from Amazon Mechanical Turk. It has three distinct domains: calendar scheduling, weather information retrieval, and point-of-interest navigation. This dataset has shorter conversation turns, but the user and system behaviors are more diverse. In addition, the system responses are variant and the KB information is much more complicated. Hence, this dataset requires stronger ability to interact with KBs, rather than dialog state tracking. 

\subsection{Training}
We trained our model end-to-end using Adam optimizer~\cite{KingmaB14}, and chose learning rate between $[1e^{-3},1e^{-4}]$. The MemNNs, both encoder and decoder, have hops $K= 1,3, 6$ to show the performance difference. We use simple greedy search and without any re-scoring techniques. The embedding size, which is also equivalent to the memory size and the RNN hidden size (i.e., including the baselines), has been selected between $[64,512]$. The dropout rate is set between $[0.1, 0.4]$, and we also randomly mask some input words into unknown tokens to simulate OOV situation with the same dropout ratio. In all the datasets, we tuned the hyper-parameters with grid-search over the validation set, using as measure to the Per-response Accuracy for bAbI dialog and DSTC2, and BLEU score for the In-Car Assistant. 

\subsection{Evaluation Metrics}
\textbf{Per-response/dialog Accuracy}: 
A generative response is correct only if it is exactly the same as the gold response. A dialog is correct only if every generated responses of the dialog are correct, which can be considered as the task-completion rate. Note that \citet{bordes2016learning} tests their model by selecting the system response from predefined response candidates, that is, their system solves a multi-class classification task. Since Mem2Seq generates each token individually, evaluating with this metric is much more challenging for our model.

\noindent\textbf{BLEU}: It is a measure commonly used for machine translation systems~\cite{papineniBLEU2002}, but it has also been used in evaluating dialog systems~\cite{eric-manning:2017:EACLshort,zhao2017generative} and chat-bots~\cite{ritter2011data,li-EtAl:2016:N16-11}. Moreover, BLEU score is a relevant measure in task-oriented dialog as there is not a large variance between the generated answers, unlike open domain generation~\cite{liu-EtAl:2016:EMNLP20163}. Hence, we include BLEU score in our evaluation (i.e. using Moses \texttt{multi-bleu.perl} script).

\noindent\textbf{Entity F1}:
We micro-average over the entire set of system responses and compare the entities in plain text. The entities in each gold system response are selected by a predefined entity list. This metric evaluates the ability to generate relevant entities from the provided KBs and to capture the semantics of the dialog~\cite{eric-manning:2017:EACLshort,ericKVR2017}. Note that the original In-Car Assistant F1 scores reported in~\citet{ericKVR2017} uses the entities in their canonicalized forms, which are not calculated based on real entity value. Since the datasets are not designed for slot-tracking, we report entity F1 rather than the slot-tracking accuracy as in~\cite{wen2016network,zhao2017generative}. 

\section{Experimental Results}
We mainly compare Mem2Seq with hop 1,3,6 with several existing models: query-reduction networks (QRN, ~\citet{seo2016query}), end-to-end memory networks (MemNN,~\citet{sukhbaatar2015end}), and gated end-to-end memory networks (GMemNN,~\citet{perez2016gated}). We also implemented the following baseline models: standard sequence-to-sequence (Seq2Seq) models with and without attention~\cite{luong-pham-manning:2015:EMNLP}, and pointer to unknown (Ptr-Unk, ~\citet{gulcehre-EtAl:2016:P16-1}). Note that the results we listed in Table~\ref{TB:babi-dialog} and Table~\ref{TB:DSTC2} for QRN are different from the original paper, because based on their released code,~\footnote{We simply modified the evaluation part and reported the results. (https://github.com/uwnlp/qrn)} 
we discovered that the per-response accuracy was not correctly computed. 

\noindent\textbf{bAbI Dialog}: 
In Table~\ref{TB:babi-dialog}, we follow~\citet{bordes2016learning} to compare the performance based on per-response and per-dialog accuracy. Mem2Seq with 6 hops can achieve per-response 97.9\% and per-dialog 69.6\% accuracy in T5, and 84.5\% and 2.3\% for T5-OOV, which surpass existing methods by far. One can find that in T3 especially, which is the task to recommend restaurant based on their ranks, our model can achieve promising results due to the memory pointer. In terms of per-response accuracy, this indicates that our model can generalize well with few performance loss for test OOV data, while others have around 15-20\% drop. The performance gain in OOV data is also mainly attributed to the use of copy mechanism. In addition, the effectiveness of hops is demonstrated in tasks 3-5, since they require reasoning ability over the KB information. 
Note that QRN, MemNN and GMemNN viewed bAbI dialog tasks as classification problems. Although their tasks are easier compared to our generative methods, Mem2Seq models can still overpass the performance.
Finally, one can find that Seq2Seq and Ptr-Unk models are also strong baselines, which further confirms that generative methods can also achieve good performance in task-oriented dialog systems~\cite{eric-manning:2017:EACLshort}.    

\noindent\textbf{DSTC2}: 
In Table~\ref{TB:DSTC2}, the Seq2Seq models from~\citet{eric-manning:2017:EACLshort} and the rule-based from~\citet{bordes2016learning} are reported. Mem2Seq has the highest 75.3\% entity F1 score and an high of 55.3 BLEU score. This further confirms that Mem2Seq can perform well in retrieving the correct entity, using the multiple hop mechanism without losing language modeling. Here, we do not report the results using match type~\cite{bordes2016learning} or entity type~\cite{eric-manning:2017:EACLshort} feature, since this meta-information are not commonly available and we want to have an evaluation on plain input output couples. One can also find out that, Mem2Seq comparable per-response accuracy (i.e. 2\% margin) among other existing solution. Note that the per-response accuracy for every model is less than 50\% since the dataset is quite noisy and it is hard to generate a response that is exactly the same as the gold one. 

\noindent\textbf{In-Car Assistant}: 
In Table~\ref{TB:INCAR}, our model can achieve highest 12.6 BLEU score. In addition, Mem2Seq has shown promising results in terms of Entity F1 scores (33.4\%), which are, in general, much higher than those of other baselines. Note that the numbers reported from~\citet{ericKVR2017} are not directly comparable to ours as we mention below. The other baselines such as Seq2Seq or Ptr-Unk especially have worse performances in this dataset since it is very inefficient for RNN methods to encode longer KB information, which is the advantage of Mem2Seq.

Furthermore, we observe an interesting phenomenon that humans can easily achieve a high entity F1 score with a low BLEU score. This implies that stronger reasoning ability over entities (hops) is crucial, but the results may not be similar to the golden answer. We believe humans can produce good answers even with a low BLEU score, since there could be different ways to express the same concepts. Therefore, Mem2Seq shows the potential to successfully choose the correct entities. 

Note that the results of KV Retrieval Net baseline reported in Table~\ref{TB:INCAR} come from the original paper~\cite{ericKVR2017} of In-Car Assistant, where they simplified the task by mapping the expression of entities to a canonical form using named entity recognition (NER) and linking. Hence the evaluation is not directly comparable to our system. For example, their model learned to generate responses such as ``You have a football game at football\_time with football\_party,'' instead of generating a sentence such as ``You have a football game at 7 pm with John.'' Since there could be more than one football\_party or football\_time, their model does not learn how to access the KBs, but it rather learns the canonicalized language model. 

\begin{figure}[t]
\includegraphics[width=\linewidth]{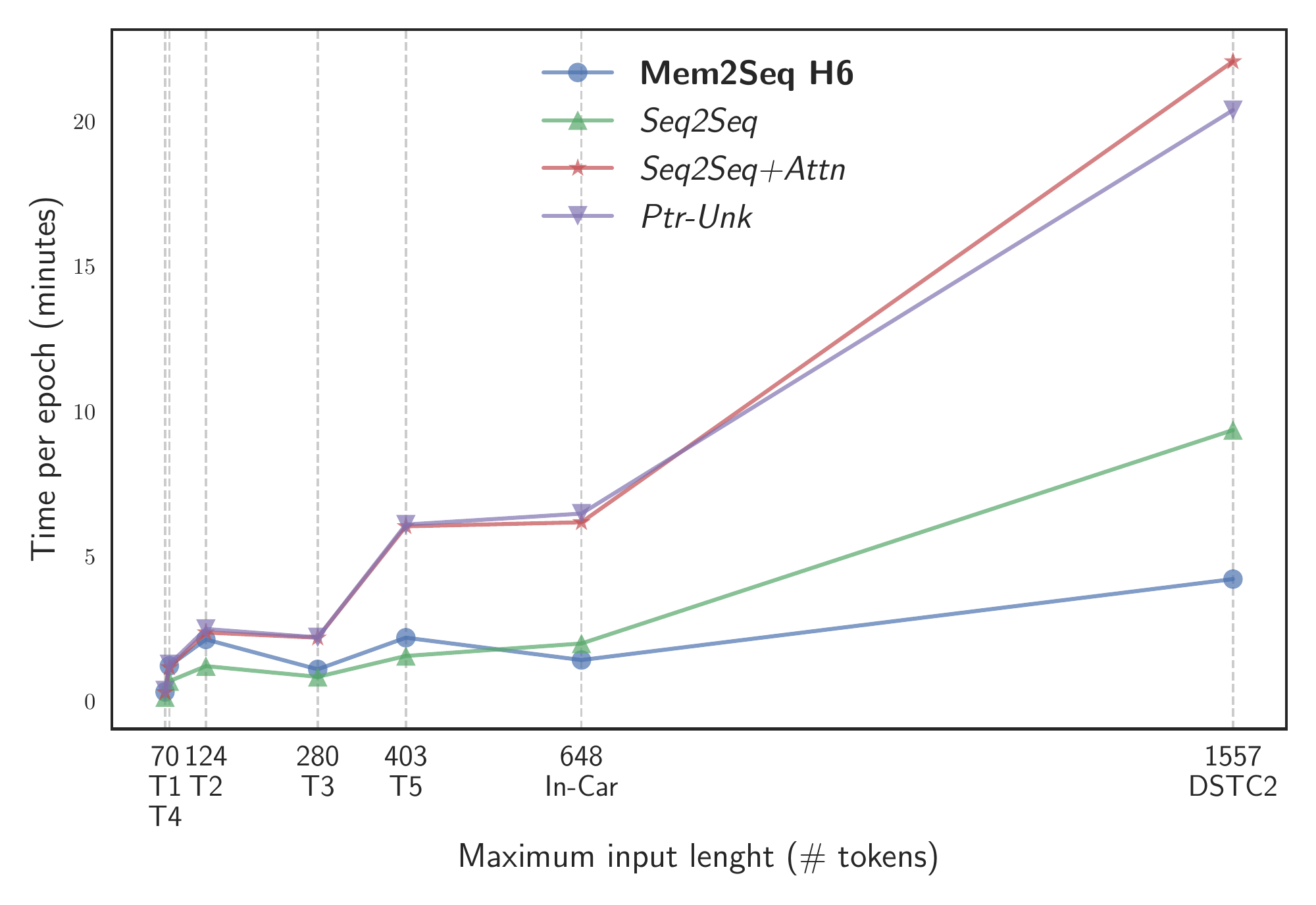}
\setlength{\abovecaptionskip}{-10pt} 
\caption{Training time per-epoch for different tasks (lower is better). The speed difference becomes larger as the maximal input length increases.}
\label{FIG:speed}
\end{figure}

\noindent\textbf{Time Per-Epoch}: 
We also compare the training time~\footnote{Intel(R) Core(TM) i7-3930K CPU@3.20GHz, using a GeForce GTX 1080 Ti} in Figure~\ref{FIG:speed}. The experiments are set with batch size 16, and we report each model with the hyper-parameter that can achieved the highest performance. One can observe that the training time is not that different for short input length (bAbI dialog tasks 1-4) and the gap becomes larger as the maximal input length increases. Mem2Seq is around 5 times faster in In-Car Assistant and DSTC2 compared to Seq2Seq with attention. This difference in training efficiency is mainly attributed to the fact that Seq2Seq models have input sequential dependencies which limit any parallelization. Moreover, it is unavoidable for Seq2Seq models to encode KBs, instead Mem2Seq only encodes with dialog history.

\begin{figure}[t]
\includegraphics[width=\linewidth]{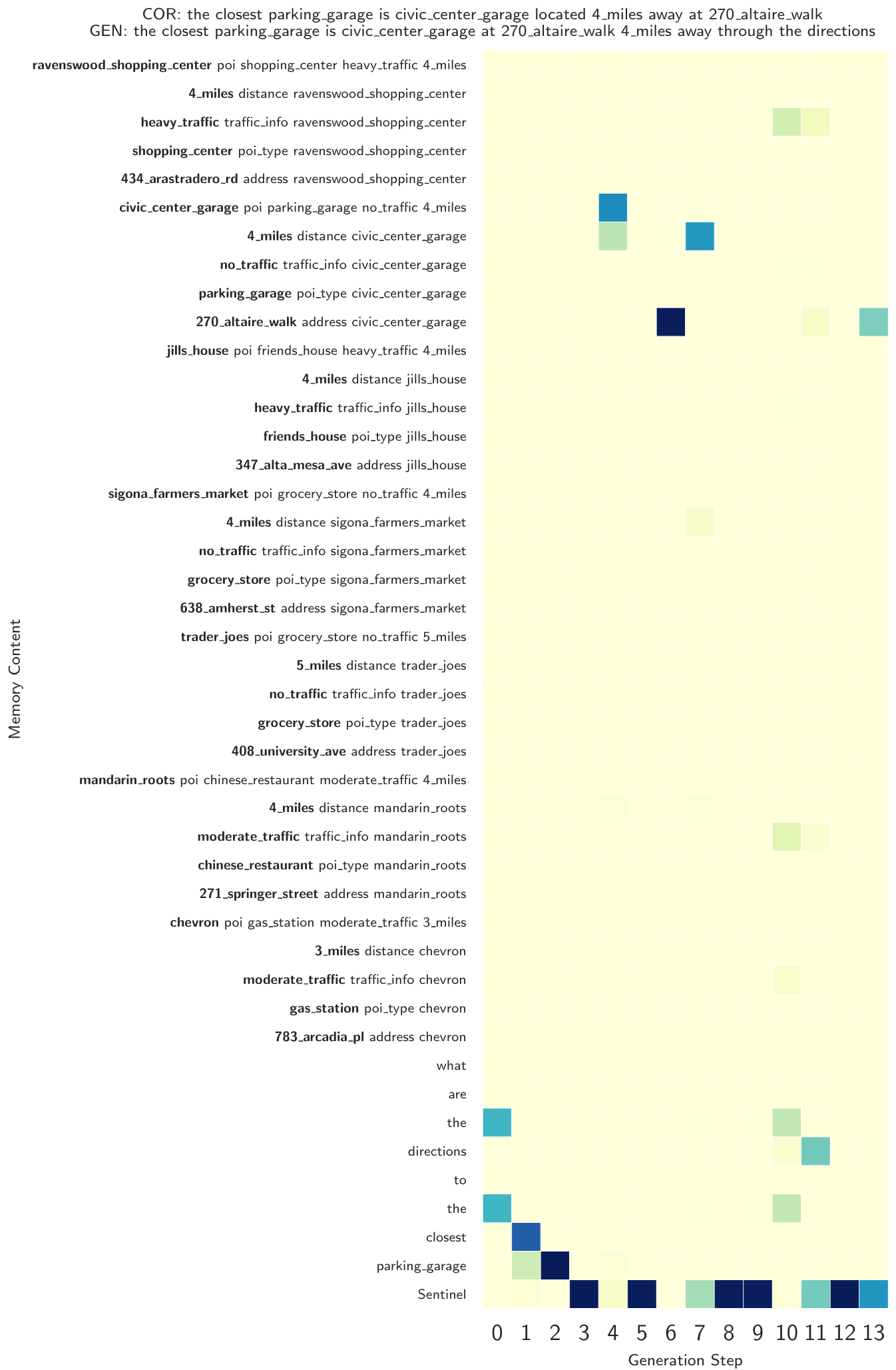}
\setlength{\abovecaptionskip}{-5pt} 
\caption{Last hop memory attention visualization from the In-Car dataset. COR and GEN on the top are the correct response and our generated one.}
\label{MEMATT}
\end{figure}

\begin{figure}[t]
\centering
\includegraphics[width=\linewidth]{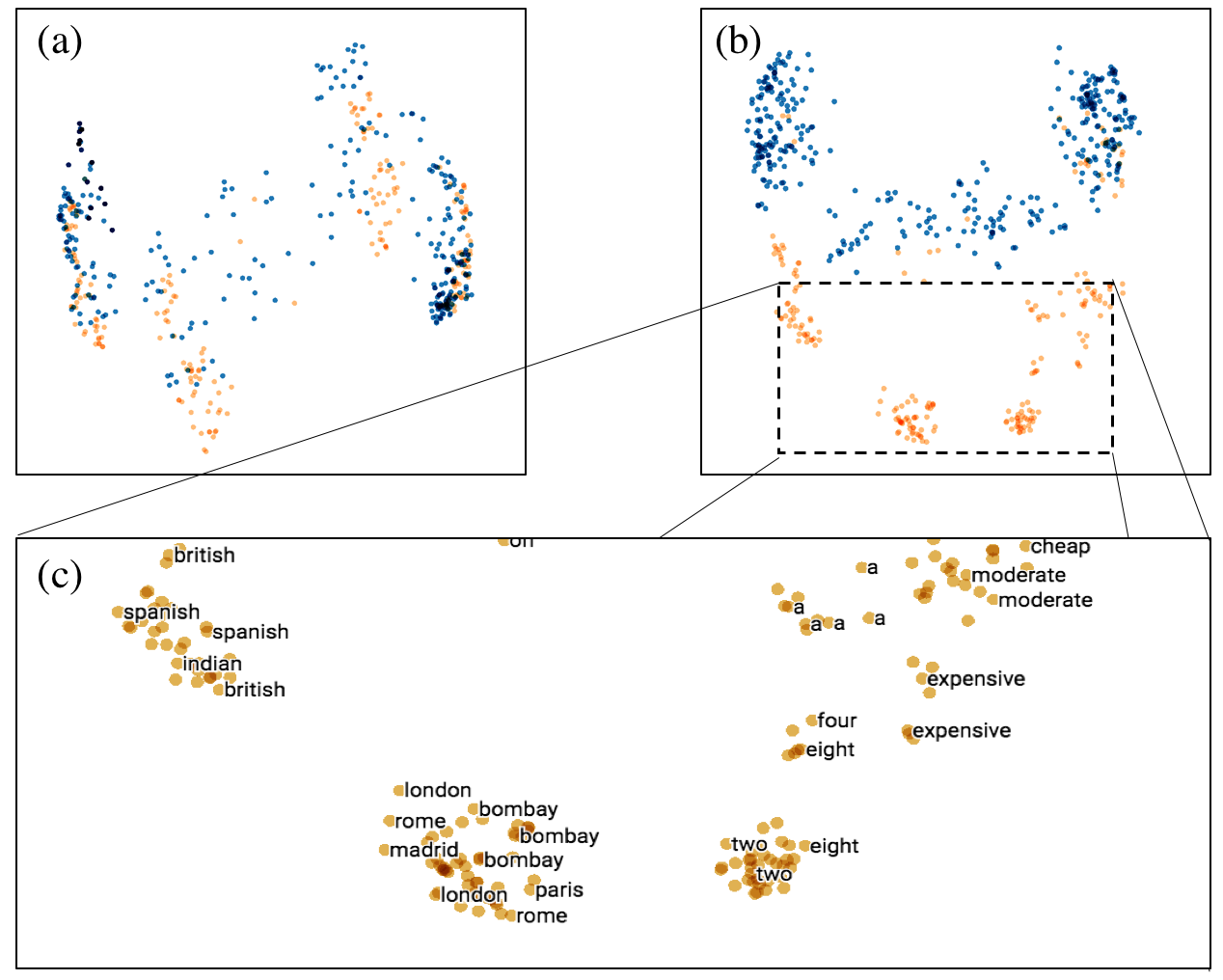}
\setlength{\abovecaptionskip}{-10pt} 
\caption{Principal component analysis of query vectors in hop (a) 1 and (b) 6 for bAbI dialog.}
\label{FIG:hopsQuery}
\end{figure}

\section{Analysis and Discussion}
\noindent\textbf{Memory Attention:} Analyzing the attention weights has been frequently used to show the memory read-out, since it is an intuitive way to understand the model dynamics. Figure~\ref{MEMATT} shows the attention vector at the last hop for each generated token. Each column represents the $P_{ptr}$ vector at the corresponding generation step. Our model has a sharp distribution over the memory, which implies that it is able to select the right token from the memory. For example, the KB information ``270\_altarie\_walk'' was retrieved at the sixth step, which is an address for ``civic\_center\_garage''. On the other hand, if the sentinel is triggered, then the generated word comes from vocabulary distribution $P_{vocab}$. For instance, the third generation step triggered the sentinel, and ``is'' is generated from the vocabulary as the word is not present in the dialog history. 

\noindent\textbf{Multiple Hops:} Mem2Seq shows how multiple hops improve the model performance in several datasets. Task 3 in the bAbI dialog dataset serves as an example, in which the systems need to recommend restaurants to users based on restaurant ranking from highest to lowest. Users can reject the recommendation and the system has to reason over the next highest restaurant. We found out there are two common patterns between hops among different samples: 1) the first hop is usually used to score all the relevant memories and retrieve information; 2) the last hop tends to focus on a specific token and makes mistakes when the attention is not sharp. Such mistakes can be attributed to lack of hops, for some samples. For more information, we report two figures in the supplementary material. 

\noindent\textbf{Query Vectors:} In Figure~\ref{FIG:hopsQuery}, the principal component analysis of Mem2Seq queries vectors is shown for different hops. Each dot is a query vector $h_t$ during each decoding time step, and it has its corresponding generated word $y_t$. The blue dots are the words generated from $P_{vocab}$, which triggered the sentinel, and orange ones are from $P_{ptr}$. One can find that in (a) hop 1, there is no clear separation of two different colors but each of which tends to group together. On the other hand, the separation becomes clearer in (b) hop 6 as each color clusters into several groups such as location, cuisine, and number. Our model tends to retrieve more information in the first hop, and points into the memories in the last hop.

\begin{table}[t]
\centering
\resizebox{\linewidth}{!}{\textit{}
\begin{tabular}{|c|c|c|c|c|c}
\hline
\textbf{room} & \textbf{time} & \textbf{date} & \textbf{party} & \textbf{event} \\ \hline
conference room 100 & 3pm & Thursday & sales team & conference \\ \hline
 - & 11am & Tuesday & - & tennis activity \\ \hline
- & 7pm & Wednesday & - & dentist appointment \\ \hline
conference room 100 &  1pm & Tuesday & HR & meeting \\ \hline
- & 7pm & Friday & - & doctor appointment \\ \hline
- &  11am & Tuesday & - & lab appointment \\ \hline
\end{tabular}
}
\resizebox{\linewidth}{!}{
\begin{tabular}{r|l}
\hline
\textbf{DRIVER}      & When's my tennis match?                    \\ \hline
\textit{Seq2seq}     & Your lab appointment is on Monday at       \\
\textit{+Att}        & Your lab appointment is on Tuesday at 11am \\
\textit{Ptr-Unk}      & Your tennis match is on PAD at 1pm         \\ \hline
\textit{Mem2Seq H1}  & Your is very welcome                       \\
\textit{Mem2Seq H3}  & Your tennis is on Tuesday at 11am   \\
\textit{Mem2Seq H6}  & Your tennis is Monday at 11am              \\\hline
\textbf{GOLD}        & Your tennis match is Tuesday at 11am      \\  \hline
\end{tabular}}
\setlength{\abovecaptionskip}{-3pt} 
\caption{Example of generated responses for the In-Car Assistant on the scheduling domain.}
\label{EXAMPL_DIALOG_2}
\end{table}

\noindent\textbf{Examples:} Table~\ref{EXAMPL_DIALOG_1} and \ref{EXAMPL_DIALOG_2} show the generated responses of different models in the two test set samples from the In-Car Assistant dataset. We report examples from this dataset since their answers are more human-like and not as structured and repetitive as others. 
Seq2Seq generally cannot produce related information, and sometimes fail in language modeling. Instead, using attention helps with this issue, but it still rarely produces the correct entities. For example, Seq2Seq with attention generated 5 miles in Table~\ref{EXAMPL_DIALOG_1} but the correct one is 4 miles. In addition, Ptr-Unk often cannot copy the correct token from the input, as shown by ``PAD'' in Table~\ref{EXAMPL_DIALOG_1}. On the other hand, Mem2Seq is able to produce the correct responses in this two examples. In particular in the navigation domain, shown in Table~\ref{EXAMPL_DIALOG_1}, Mem2Seq produces a different but still correct utterance. We report further examples from all the domains in the supplementary material. 

\noindent\textbf{Discussions:} Conventional task-oriented dialog systems~\cite{williams2007partially}, which are still widely used in commercial systems, require a multitude of human efforts in system designing and data collection. On the other hand, although end-to-end dialog systems are not perfect yet, they require much less human interference, especially in the dataset construction, as raw conversational text and KB information can be used directly without the need of heavy preprocessing (e.g. NER, dependency parsing). To this extent, Mem2Seq is a simple generative model that is able to incorporate KB information with promising generalization ability. We also discovered that the entity F1 score may be a more comprehensive evaluation metric than per-response accuracy or BLEU score, as humans can normally choose the right entities but have very diversified responses. Indeed, we want to highlight that humans may have a low BLEU score despite their correctness because there may not be a large n-gram overlap between the given response and the expected one. However, this does not imply that there is no correlation between BLEU score and human evaluation. In fact, unlike chat-bots and open domain dialogs where BLEU score does not correlate with human evaluation~\cite{liu-EtAl:2016:EMNLP20163}, in task-oriented dialogs the answers are constrained to particular entities and recurrent patterns. Thus, we believe BLEU score still can be considered as a relevant measure. In future works, several methods could be applied (e.g. Reinforcement Learning~\cite{ranzato2015sequence}, Beam Search~\cite{wiseman-rush:2016:EMNLP2016}) to improve both responses relevance and entity F1 score. However, we preferred to keep our model as simple as possible in order to show that it works well even without advanced training methods.

\section{Related Works}
End-to-end task-oriented dialog systems train a single model directly on text transcripts of dialogs~\cite{wen2016network,serban2016building,williams2017hybrid,zhao2017generative, seo2016query,serban2017hierarchical}. Here, RNNs play an important role due to their ability to create a latent representation, avoiding the need for artificial state labels. End-to-End Memory Networks~\cite{bordes2016learning,sukhbaatar2015end}, and its variants~\cite{perez2016gated,wu2017dstc6,dqmem} have also shown good results in such tasks. In each of these architectures, the output is produced by generating a sequence of tokens, or by selecting a set of predefined utterances.

Sequence-to-sequence (Seq2Seq) models have also been used in task-oriented dialog systems~\cite{zhao2017generative}. These architectures have better language modeling ability, but they do not work well in KB retrieval. Even with sophisticated attention models~\cite{luong-pham-manning:2015:EMNLP,bahdanau2014neural}, Seq2Seq fails to map the correct entities to the generated input. To alleviate this problem, copy augmented Seq2Seq models~\citet{eric-manning:2017:EACLshort}, were used. These models outperform utterance selection methods by copying relevant information directly from the KBs. Copy mechanisms has also been used in question answering tasks~\cite{Dehghani:2017:LAC:3132847.3133010,he-EtAl:2017:Long1}, neural machine translation~\cite{gulcehre-EtAl:2016:P16-1,gu-EtAl:2016:P16-1}, language modeling~\cite{merity2016pointer}, and summarization~\cite{see-liu-manning:2017:Long}. 

Less related to dialog systems, but related to our work, are the memory based decoders and the non-recurrent generative models: 1) Mem2Seq query generation phase used to access our memories can be seen as the memory controller used in Memory Augmented Neural Networks (MANN)~\cite{graves2014neural,graves2016hybrid}. Similarly, memory encoders have been used in neural machine translation~\cite{wang-EtAl:2016:EMNLP20161}, and meta-learning application~\cite{DBLP:journals/corr/KaiserNRB17}. However, Mem2Seq differs from these models as such: it uses multi-hop attention in combination with copy mechanism, whereas other models use a single matrix representation. 2) non-recurrent generative models~\cite{vaswani2017attention}, which only rely on self-attention mechanism, are related to the multi-hop attention mechanism used in MemNN.

\section{Conclusion}
In this work, we present an end-to-end trainable Memory-to-Sequence model for task-oriented dialog systems. Mem2Seq combines the multi-hop attention mechanism in end-to-end memory networks with the idea of pointer networks to incorporate external information. We empirically show our model's ability to produce relevant answers using both the external KB information and the predefined vocabulary, and visualize how the multi-hop attention mechanisms help in learning correlations between memories. Mem2Seq is fast, general, and able to achieve state-of-the-art results in three different datasets.

\section*{Acknowledgments}
This work is partially funded by ITS/319/16FP of Innovation Technology Commission, HKUST 16214415 \& 16248016 of Hong Kong Research Grants Council, and RDC 1718050-0 of
EMOS.AI.

\bibliography{acl2018}

\begin{thebibliography}{42}
\expandafter\ifx\csname natexlab\endcsname\relax\def\natexlab#1{#1}\fi

\bibitem[{Bahdanau et~al.(2015)Bahdanau, Cho, and Bengio}]{bahdanau2014neural}
Dzmitry Bahdanau, Kyunghyun Cho, and Yoshua Bengio. 2015.
\newblock Neural machine translation by jointly learning to align and
  translate.
\newblock \emph{International Conference on Learning Representations}.

\bibitem[{Bordes and Weston(2017)}]{bordes2016learning}
Antoine Bordes and Jason Weston. 2017.
\newblock Learning end-to-end goal-oriented dialog.
\newblock \emph{International Conference on Learning Representations},
  abs/1605.07683.

\bibitem[{Chung et~al.(2014)Chung, Gulcehre, Cho, and Bengio}]{GRU}
Junyoung Chung, Caglar Gulcehre, Kyunghyun Cho, and Yoshua Bengio. 2014.
\newblock Empirical evaluation of gated recurrent neural networks on sequence
  modeling.
\newblock \emph{NIPS Deep Learning and Representation Learning Workshop}.

\bibitem[{Dehghani et~al.(2017)Dehghani, Rothe, Alfonseca, and
  Fleury}]{Dehghani:2017:LAC:3132847.3133010}
Mostafa Dehghani, Sascha Rothe, Enrique Alfonseca, and Pascal Fleury. 2017.
\newblock \href {https://doi.org/10.1145/3132847.3133010} {Learning to attend,
  copy, and generate for session-based query suggestion}.
\newblock In \emph{Proceedings of the 2017 ACM on Conference on Information and
  Knowledge Management}, CIKM '17, pages 1747--1756, New York, NY, USA. ACM.

\bibitem[{Eric et~al.(2017)Eric, Krishnan, Charette, and Manning}]{ericKVR2017}
Mihail Eric, Lakshmi Krishnan, Francois Charette, and Christopher~D. Manning.
  2017.
\newblock \href {http://aclweb.org/anthology/W17-5506} {Key-value retrieval
  networks for task-oriented dialogue}.
\newblock In \emph{Proceedings of the 18th Annual SIGdial Meeting on Discourse
  and Dialogue}, pages 37--49. Association for Computational Linguistics.

\bibitem[{Eric and Manning(2017)}]{eric-manning:2017:EACLshort}
Mihail Eric and Christopher Manning. 2017.
\newblock \href {http://www.aclweb.org/anthology/E17-2075} {A copy-augmented
  sequence-to-sequence architecture gives good performance on task-oriented
  dialogue}.
\newblock In \emph{Proceedings of the 15th Conference of the European Chapter
  of the Association for Computational Linguistics: Volume 2, Short Papers},
  pages 468--473, Valencia, Spain. Association for Computational Linguistics.

\bibitem[{Graves et~al.(2014)Graves, Wayne, and Danihelka}]{graves2014neural}
Alex Graves, Greg Wayne, and Ivo Danihelka. 2014.
\newblock Neural turing machines.
\newblock \emph{CoRR}.

\bibitem[{Graves et~al.(2016)Graves, Wayne, Reynolds, Harley, Danihelka,
  Grabska-Barwi{\'n}ska, Colmenarejo, Grefenstette, Ramalho, Agapiou
  et~al.}]{graves2016hybrid}
Alex Graves, Greg Wayne, Malcolm Reynolds, Tim Harley, Ivo Danihelka, Agnieszka
  Grabska-Barwi{\'n}ska, Sergio~G{\'o}mez Colmenarejo, Edward Grefenstette,
  Tiago Ramalho, John Agapiou, et~al. 2016.
\newblock Hybrid computing using a neural network with dynamic external memory.
\newblock \emph{Nature}, 538(7626):471--476.

\bibitem[{Gu et~al.(2016)Gu, Lu, Li, and Li}]{gu-EtAl:2016:P16-1}
Jiatao Gu, Zhengdong Lu, Hang Li, and Victor~O.K. Li. 2016.
\newblock \href {http://www.aclweb.org/anthology/P16-1154} {Incorporating
  copying mechanism in sequence-to-sequence learning}.
\newblock In \emph{Proceedings of the 54th Annual Meeting of the Association
  for Computational Linguistics (Volume 1: Long Papers)}, pages 1631--1640,
  Berlin, Germany. Association for Computational Linguistics.

\bibitem[{Gulcehre et~al.(2016)Gulcehre, Ahn, Nallapati, Zhou, and
  Bengio}]{gulcehre-EtAl:2016:P16-1}
Caglar Gulcehre, Sungjin Ahn, Ramesh Nallapati, Bowen Zhou, and Yoshua Bengio.
  2016.
\newblock \href {http://www.aclweb.org/anthology/P16-1014} {Pointing the
  unknown words}.
\newblock In \emph{Proceedings of the 54th Annual Meeting of the Association
  for Computational Linguistics (Volume 1: Long Papers)}, pages 140--149,
  Berlin, Germany. Association for Computational Linguistics.

\bibitem[{He et~al.(2017)He, Liu, Liu, and Zhao}]{he-EtAl:2017:Long1}
Shizhu He, Cao Liu, Kang Liu, and Jun Zhao. 2017.
\newblock \href {http://aclweb.org/anthology/P17-1019} {Generating natural
  answers by incorporating copying and retrieving mechanisms in
  sequence-to-sequence learning}.
\newblock In \emph{Proceedings of the 55th Annual Meeting of the Association
  for Computational Linguistics (Volume 1: Long Papers)}, pages 199--208,
  Vancouver, Canada. Association for Computational Linguistics.

\bibitem[{Henderson et~al.(2014)Henderson, Thomson, and
  Williams}]{henderson2014second}
Matthew Henderson, Blaise Thomson, and Jason~D Williams. 2014.
\newblock The second dialog state tracking challenge.
\newblock In \emph{Proceedings of the 15th Annual Meeting of the Special
  Interest Group on Discourse and Dialogue (SIGDIAL)}, pages 263--272.

\bibitem[{Hori et~al.(2009)Hori, Ohtake, Misu, Kashioka, and
  Nakamura}]{hori2009statistical}
Chiori Hori, Kiyonori Ohtake, Teruhisa Misu, Hideki Kashioka, and Satoshi
  Nakamura. 2009.
\newblock Statistical dialog management applied to wfst-based dialog systems.
\newblock In \emph{IEEE International Conference on Acoustics, Speech and
  Signal Processing, 2009. ICASSP 2009.}, pages 4793--4796. IEEE.

\bibitem[{Kaiser et~al.(2017)Kaiser, Nachum, Roy, and
  Bengio}]{DBLP:journals/corr/KaiserNRB17}
Lukasz Kaiser, Ofir Nachum, Aurko Roy, and Samy Bengio. 2017.
\newblock Learning to remember rare events.
\newblock \emph{International Conference on Learning Representations}.

\bibitem[{Kingma and Ba(2015)}]{KingmaB14}
Diederik~P Kingma and Jimmy Ba. 2015.
\newblock Adam: A method for stochastic optimization.
\newblock \emph{International Conference on Learning Representations}.

\bibitem[{Lee et~al.(2009)Lee, Jung, Kim, and Lee}]{lee2009example}
Cheongjae Lee, Sangkeun Jung, Seokhwan Kim, and Gary~Geunbae Lee. 2009.
\newblock Example-based dialog modeling for practical multi-domain dialog
  system.
\newblock \emph{Speech Communication}, 51(5):466--484.

\bibitem[{Levin et~al.(2000)Levin, Pieraccini, and
  Eckert}]{levin2000stochastic}
Esther Levin, Roberto Pieraccini, and Wieland Eckert. 2000.
\newblock A stochastic model of human-machine interaction for learning dialog
  strategies.
\newblock \emph{IEEE Transactions on speech and audio processing}, 8(1):11--23.

\bibitem[{Li et~al.(2016)Li, Galley, Brockett, Gao, and
  Dolan}]{li-EtAl:2016:N16-11}
Jiwei Li, Michel Galley, Chris Brockett, Jianfeng Gao, and Bill Dolan. 2016.
\newblock \href {http://www.aclweb.org/anthology/N16-1014} {A
  diversity-promoting objective function for neural conversation models}.
\newblock In \emph{Proceedings of the 2016 Conference of the North American
  Chapter of the Association for Computational Linguistics: Human Language
  Technologies}, pages 110--119, San Diego, California. Association for
  Computational Linguistics.

\bibitem[{Liu et~al.(2016)Liu, Lowe, Serban, Noseworthy, Charlin, and
  Pineau}]{liu-EtAl:2016:EMNLP20163}
Chia-Wei Liu, Ryan Lowe, Iulian Serban, Mike Noseworthy, Laurent Charlin, and
  Joelle Pineau. 2016.
\newblock \href {https://aclweb.org/anthology/D16-1230} {How not to evaluate
  your dialogue system: An empirical study of unsupervised evaluation metrics
  for dialogue response generation}.
\newblock In \emph{Proceedings of the 2016 Conference on Empirical Methods in
  Natural Language Processing}, pages 2122--2132, Austin, Texas. Association
  for Computational Linguistics.

\bibitem[{Liu and Perez(2017)}]{perez2016gated}
Fei Liu and Julien Perez. 2017.
\newblock \href {http://www.aclweb.org/anthology/E17-1001} {Gated end-to-end
  memory networks}.
\newblock In \emph{Proceedings of the 15th Conference of the European Chapter
  of the Association for Computational Linguistics: Volume 1, Long Papers},
  pages 1--10, Valencia, Spain. Association for Computational Linguistics.

\bibitem[{Luong et~al.(2015)Luong, Pham, and
  Manning}]{luong-pham-manning:2015:EMNLP}
Thang Luong, Hieu Pham, and Christopher~D. Manning. 2015.
\newblock \href {http://aclweb.org/anthology/D15-1166} {Effective approaches to
  attention-based neural machine translation}.
\newblock In \emph{Proceedings of the 2015 Conference on Empirical Methods in
  Natural Language Processing}, pages 1412--1421, Lisbon, Portugal. Association
  for Computational Linguistics.

\bibitem[{Merity et~al.(2017)Merity, Xiong, Bradbury, and
  Socher}]{merity2016pointer}
Stephen Merity, Caiming Xiong, James Bradbury, and Richard Socher. 2017.
\newblock Pointer sentinel mixture models.
\newblock \emph{International Conference on Learning Representations}.

\bibitem[{Miller et~al.(2016)Miller, Fisch, Dodge, Karimi, Bordes, and
  Weston}]{miller-EtAl:2016:EMNLP2016}
Alexander Miller, Adam Fisch, Jesse Dodge, Amir-Hossein Karimi, Antoine Bordes,
  and Jason Weston. 2016.
\newblock \href {https://aclweb.org/anthology/D16-1147} {Key-value memory
  networks for directly reading documents}.
\newblock In \emph{Proceedings of the 2016 Conference on Empirical Methods in
  Natural Language Processing}, pages 1400--1409, Austin, Texas. Association
  for Computational Linguistics.

\bibitem[{Papineni et~al.(2002)Papineni, Roukos, Ward, and
  Zhu}]{papineniBLEU2002}
Kishore Papineni, Salim Roukos, Todd Ward, and Wei-Jing Zhu. 2002.
\newblock \href {https://doi.org/10.3115/1073083.1073135} {Bleu: a method for
  automatic evaluation of machine translation}.
\newblock In \emph{Proceedings of 40th Annual Meeting of the Association for
  Computational Linguistics}, pages 311--318, Philadelphia, Pennsylvania, USA.
  Association for Computational Linguistics.

\bibitem[{Ranzato et~al.(2016)Ranzato, Chopra, Auli, and
  Zaremba}]{ranzato2015sequence}
Marc'Aurelio Ranzato, Sumit Chopra, Michael Auli, and Wojciech Zaremba. 2016.
\newblock Sequence level training with recurrent neural networks.
\newblock \emph{International Conference on Learning Representations}.

\bibitem[{Ritter et~al.(2011)Ritter, Cherry, and Dolan}]{ritter2011data}
Alan Ritter, Colin Cherry, and William~B. Dolan. 2011.
\newblock \href {http://www.aclweb.org/anthology/D11-1054} {Data-driven
  response generation in social media}.
\newblock In \emph{Proceedings of the 2011 Conference on Empirical Methods in
  Natural Language Processing}, pages 583--593, Edinburgh, Scotland, UK.
  Association for Computational Linguistics.

\bibitem[{See et~al.(2017)See, Liu, and Manning}]{see-liu-manning:2017:Long}
Abigail See, Peter~J. Liu, and Christopher~D. Manning. 2017.
\newblock \href {http://aclweb.org/anthology/P17-1099} {Get to the point:
  Summarization with pointer-generator networks}.
\newblock In \emph{Proceedings of the 55th Annual Meeting of the Association
  for Computational Linguistics (Volume 1: Long Papers)}, pages 1073--1083,
  Vancouver, Canada. Association for Computational Linguistics.

\bibitem[{Seo et~al.(2017)Seo, Min, Farhadi, and Hajishirzi}]{seo2016query}
Minjoon Seo, Sewon Min, Ali Farhadi, and Hannaneh Hajishirzi. 2017.
\newblock Query-reduction networks for question answering.
\newblock \emph{International Conference on Learning Representations}.

\bibitem[{Serban et~al.(2016)Serban, Sordoni, Bengio, Courville, and
  Pineau}]{serban2016building}
Iulian~Vlad Serban, Alessandro Sordoni, Yoshua Bengio, Aaron~C Courville, and
  Joelle Pineau. 2016.
\newblock Building end-to-end dialogue systems using generative hierarchical
  neural network models.
\newblock In \emph{AAAI}, pages 3776--3784.

\bibitem[{Serban et~al.(2017)Serban, Sordoni, Lowe, Charlin, Pineau, Courville,
  and Bengio}]{serban2017hierarchical}
Iulian~Vlad Serban, Alessandro Sordoni, Ryan Lowe, Laurent Charlin, Joelle
  Pineau, Aaron~C Courville, and Yoshua Bengio. 2017.
\newblock A hierarchical latent variable encoder-decoder model for generating
  dialogues.
\newblock In \emph{AAAI}, pages 3295--3301.

\bibitem[{Sukhbaatar et~al.(2015)Sukhbaatar, Weston, Fergus
  et~al.}]{sukhbaatar2015end}
Sainbayar Sukhbaatar, Jason Weston, Rob Fergus, et~al. 2015.
\newblock End-to-end memory networks.
\newblock In \emph{Advances in neural information processing systems}, pages
  2440--2448.

\bibitem[{Vaswani et~al.(2017)Vaswani, Shazeer, Parmar, Uszkoreit, Jones,
  Gomez, Kaiser, and Polosukhin}]{vaswani2017attention}
Ashish Vaswani, Noam Shazeer, Niki Parmar, Jakob Uszkoreit, Llion Jones,
  Aidan~N Gomez, {\L}ukasz Kaiser, and Illia Polosukhin. 2017.
\newblock Attention is all you need.
\newblock In \emph{Advances in Neural Information Processing Systems}, pages
  6000--6010.

\bibitem[{Vinyals et~al.(2015)Vinyals, Fortunato, and Jaitly}]{NIPS2015_5866}
Oriol Vinyals, Meire Fortunato, and Navdeep Jaitly. 2015.
\newblock \href {http://papers.nips.cc/paper/5866-pointer-networks.pdf}
  {Pointer networks}.
\newblock In C.~Cortes, N.~D. Lawrence, D.~D. Lee, M.~Sugiyama, and R.~Garnett,
  editors, \emph{Advances in Neural Information Processing Systems 28}, pages
  2692--2700. Curran Associates, Inc.

\bibitem[{Wang et~al.(2016)Wang, Lu, Li, and Liu}]{wang-EtAl:2016:EMNLP20161}
Mingxuan Wang, Zhengdong Lu, Hang Li, and Qun Liu. 2016.
\newblock \href {https://aclweb.org/anthology/D16-1027} {Memory-enhanced
  decoder for neural machine translation}.
\newblock In \emph{Proceedings of the 2016 Conference on Empirical Methods in
  Natural Language Processing}, pages 278--286, Austin, Texas. Association for
  Computational Linguistics.

\bibitem[{Wen et~al.(2017)Wen, Gasic, Mrksic, Rojas-Barahona, hao Su, Ultes,
  Vandyke, and Young}]{wen2016network}
Tsung-Hsien Wen, Milica Gasic, Nikola Mrksic, Lina~Maria Rojas-Barahona, Pei
  hao Su, Stefan Ultes, David Vandyke, and Steve~J. Young. 2017.
\newblock A network-based end-to-end trainable task-oriented dialogue system.
\newblock In \emph{EACL}.

\bibitem[{Williams et~al.(2017)Williams, Asadi, and Zweig}]{williams2017hybrid}
Jason~D Williams, Kavosh Asadi, and Geoffrey Zweig. 2017.
\newblock \href {http://aclweb.org/anthology/P17-1062} {Hybrid code networks:
  practical and efficient end-to-end dialog control with supervised and
  reinforcement learning}.
\newblock In \emph{Proceedings of the 55th Annual Meeting of the Association
  for Computational Linguistics (Volume 1: Long Papers)}, pages 665--677,
  Vancouver, Canada. Association for Computational Linguistics.

\bibitem[{Williams and Young(2007)}]{williams2007partially}
Jason~D Williams and Steve Young. 2007.
\newblock Partially observable markov decision processes for spoken dialog
  systems.
\newblock \emph{Computer Speech \& Language}, 21(2):393--422.

\bibitem[{Wiseman and Rush(2016)}]{wiseman-rush:2016:EMNLP2016}
Sam Wiseman and Alexander~M. Rush. 2016.
\newblock \href {https://aclweb.org/anthology/D16-1137} {Sequence-to-sequence
  learning as beam-search optimization}.
\newblock In \emph{Proceedings of the 2016 Conference on Empirical Methods in
  Natural Language Processing}, pages 1296--1306, Austin, Texas. Association
  for Computational Linguistics.

\bibitem[{Wu et~al.(2017)Wu, Madotto, Winata, and Fung}]{wu2017dstc6}
Chien-Sheng Wu, Andrea Madotto, Genta Winata, and Pascale Fung. 2017.
\newblock End-to-end recurrent entity network for entity-value independent
  goal-oriented dialog learning.
\newblock In \emph{Dialog System Technology Challenges Workshop, DSTC6}.

\bibitem[{Wu et~al.(2018)Wu, Madotto, Winata, and Fung}]{dqmem}
Chien-Sheng Wu, Andrea Madotto, Genta Winata, and Pascale Fung. 2018.
\newblock End-to-end dynamic query memory network for entity-value independent
  task-oriented dialog.
\newblock In \emph{IEEE International Conference on Acoustics, Speech and
  Signal Processing (ICASSP)}.

\bibitem[{Young et~al.(2013)Young, Ga{\v{s}}i{\'c}, Thomson, and
  Williams}]{young2013pomdp}
Steve Young, Milica Ga{\v{s}}i{\'c}, Blaise Thomson, and Jason~D Williams.
  2013.
\newblock Pomdp-based statistical spoken dialog systems: A review.
\newblock \emph{Proceedings of the IEEE}, 101(5):1160--1179.

\bibitem[{Zhao et~al.(2017)Zhao, Lu, Lee, and Eskenazi}]{zhao2017generative}
Tiancheng Zhao, Allen Lu, Kyusong Lee, and Maxine Eskenazi. 2017.
\newblock \href {http://aclweb.org/anthology/W17-5505} {Generative
  encoder-decoder models for task-oriented spoken dialog systems with chatting
  capability}.
\newblock In \emph{Proceedings of the 18th Annual SIGdial Meeting on Discourse
  and Dialogue}, pages 27--36. Association for Computational Linguistics.

\end{thebibliography}
\bibliographystyle{acl_natbib}

\onecolumn

\section{Tables}
\subsection{Time per Epoch}
\begin{table}[H]
\centering
\resizebox{\linewidth}{!}{
\begin{tabular}{|r|c|c|c|c|c|c|c|}
\hline
\multicolumn{1}{|c|}{\textit{}} & \textbf{T1} & \textbf{T2} & \textbf{T3} & \textbf{T4} & \textbf{T5} & \textbf{DSTC2} & \textbf{In-Car} \\ \hline
\textit{Seq2Seq}                & 0.7         & 1.22        & 0.85        & 0.15        & 1.57        & 9.36           & 2.00            \\ \hline
\textit{+Attn}                  & 1.18        & 2.38        & 2.2         & 0.32        & 6.04        & 22.07          & 6.18            \\ \hline
\textit{Ptr-Unk}                & 1.28        & 2.5         & 2.21        & 0.38        & 6.10        & 20.39          & 6.48            \\ \hline
\textit{\textbf{Mem2Seq H1}}    & 0.73        & 1.10        & 0.55        & 0.15        & 1.05        & 1.49           & 0.68            \\ \hline
\textit{\textbf{Mem2Seq H3}}    & 0.97        & 1.48        & 0.77        & 0.21        & 1.33        & 2.36           & 1.02            \\ \hline
\textit{\textbf{Mem2Seq H6}}    & 1.23        & 2.14        & 1.11        & 0.33        & 2.20        & 4.22           & 1.43            \\ \hline
\end{tabular}}
\setlength{\abovecaptionskip}{-4pt} 
\caption{Minutes per-epoch. Mem2Seq is faster compared to others especially for longer inputs.}
\label{TB:TIME}
\end{table}

\subsection{Hyper-parameters}
\begin{table}[H]
\centering
\resizebox{\linewidth}{!}{
\begin{tabular}{|r|c|c|c|c|c|c|c|c|}
\hline
\multicolumn{1}{|l|}{}  & \textbf{T1} & \textbf{T2} & \textbf{T3} & \textbf{T4} & \textbf{T5} & \textbf{DSTC2} & \textbf{In-Car}      & \textit{lr}             \\ \hline
\textit{Seq2Seq}        & 256 (0.1)   & 256 (0.1)   & 128(0.1)    & 256 (0.1)   & 256 (0.1)   & -              & 512 (0.3)            & \multirow{3}{*}{0.0001} \\ \cline{1-8}
\textit{+Attn}          & 256 (0.1)   & 256 (0.1)   & 256 (0.1)   & 256 (0.1)   & 128 (0.1)   & -              & 512 (0.0)            &                         \\ \cline{1-8}
\textit{Ptr-Unk}        & 256 (0.1)   & 256 (0.1)   & 256 (0.1)   & 256 (0.1)   & 128 (0.1)   & -              & 512 (0.0)            &                         \\ \hline
\textit{Mem2Seq GRU L1} & 256 (0.3)   & 256 (0.2)   & 128 (0.2)   & 256 (0.2)   & 256 (0.1)   & 256 (0.1)      & 256 (0.2)            & \multirow{3}{*}{0.001}  \\ \cline{1-8}
\textit{Mem2Seq GRU L3} & 256 (0.1)    & 256 (0.3)   & 128 (0.2)   & 256 (0.3)    & 128 (0.1)   & 128 (0.2)    & 256 (0.1)            &                         \\ \cline{1-8}
\textit{Mem2Seq GRU L6} & 256 (0.2)    & 128 (0.3)   & 256 (0.1)   & 128 (0.3)    & 256 (0.2)   & 256 (0.1)    & 256 (0.2)			  &                         \\ \hline
\end{tabular}}
\caption{Selected hyper-parameters in each datasets for different hops. The values is the embedding dimension and the GRU hidden size, and the values between parenthesis is the dropout rate. For all the models we used learning rate equal to 0.001, with a decay rate of 0.5. }
\end{table}

\section{Visualization}
\subsection{Multiple Hops Attention}

\begin{figure}[H]
\centering
\includegraphics[width=\linewidth]{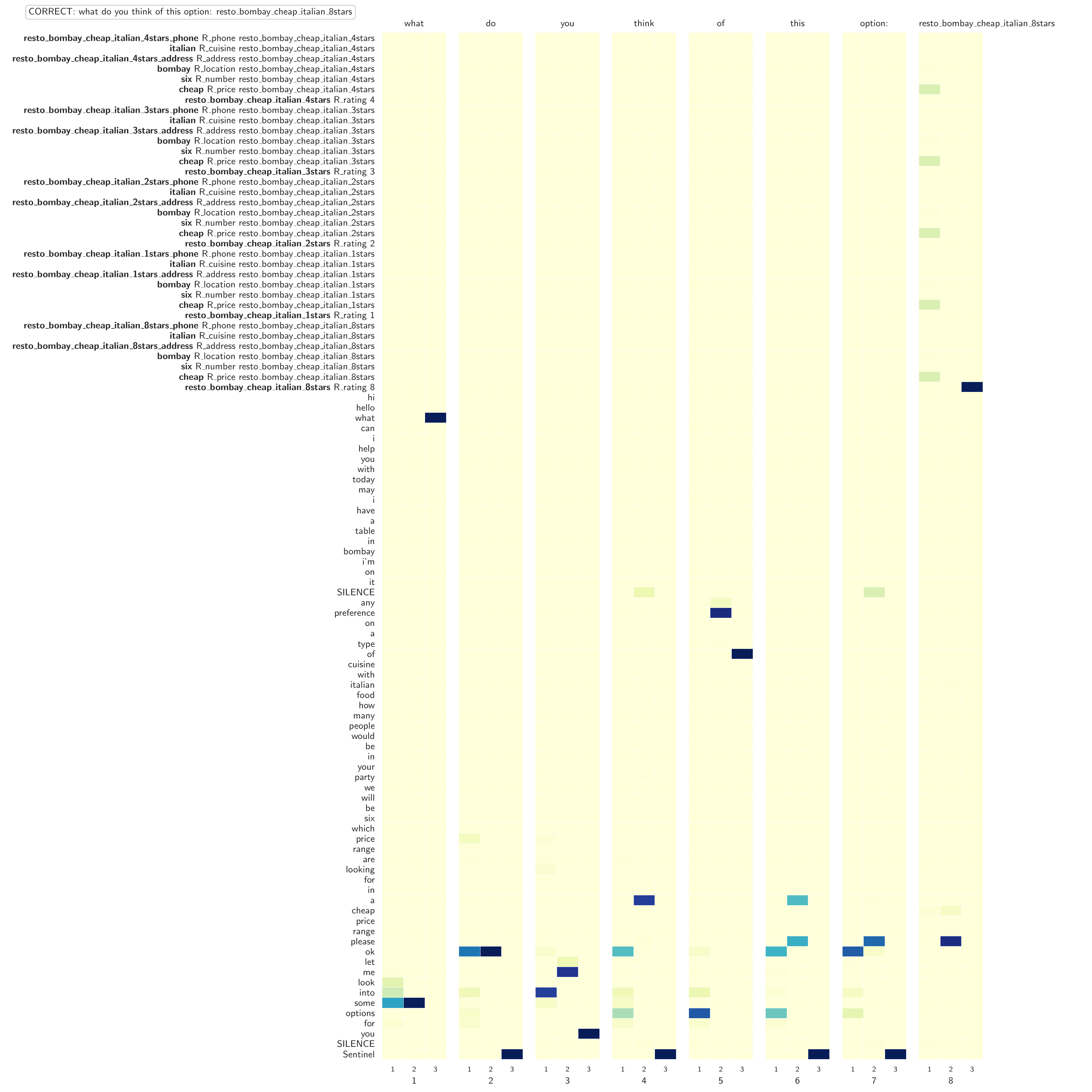}
\caption{3-hop memory attention visualization from bAbI dialog task 3. In the generation step 8, the attention at hop 1 is shallow and divided equally over the 5 possible choice, and becomes very sharp at hop 3.}
\label{3hop}
\end{figure}

\begin{figure}[H]
\centering
\includegraphics[width=\linewidth]{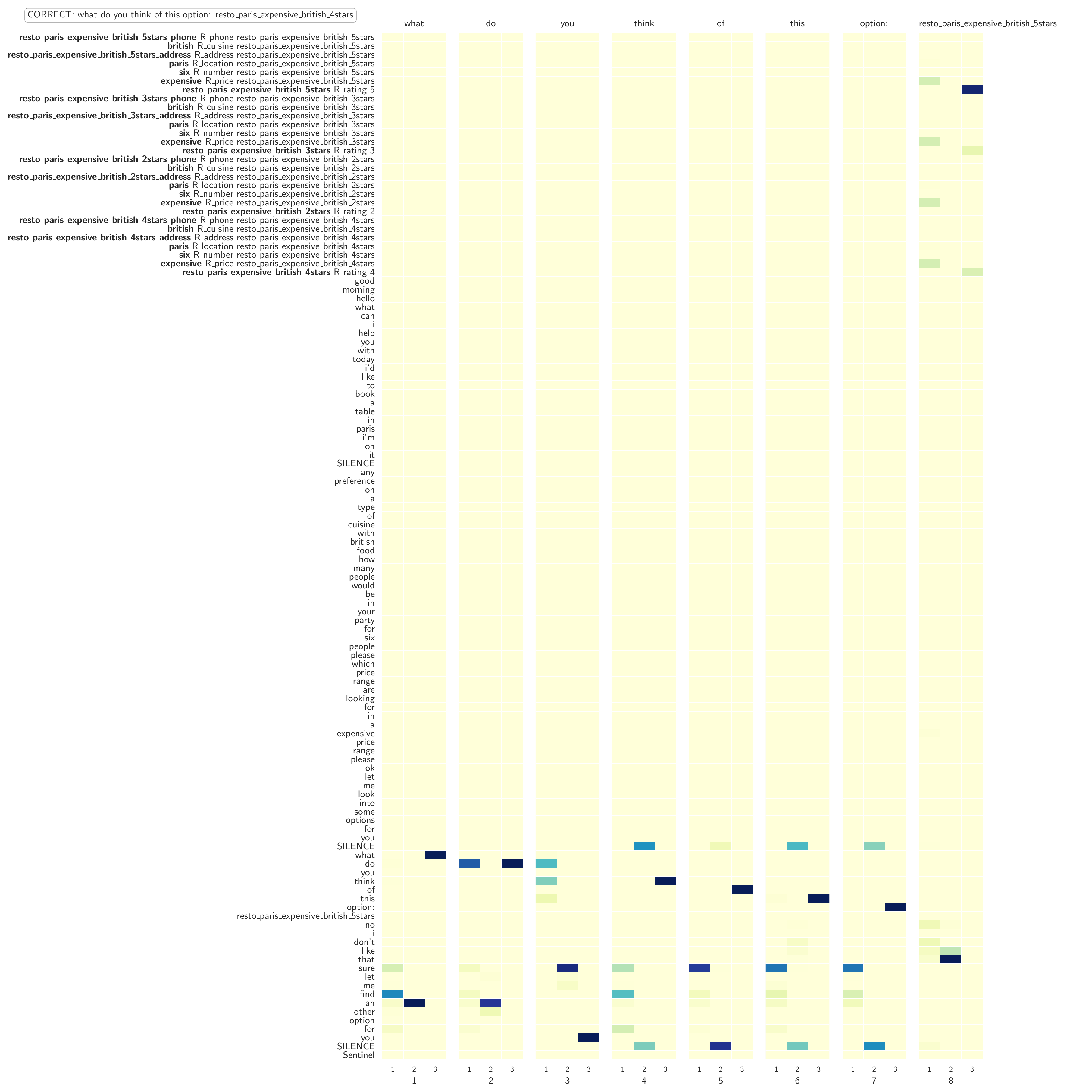}
\caption{3-hop memory attention visualization from bAbI dialog task 3. The model tends to make mistake if the attention at last hop is not sharp.}
\label{3hopWRONG}
\end{figure}

\subsection{Last Hop Attention}
\begin{figure}[H]
\centering
\includegraphics[scale=0.7]{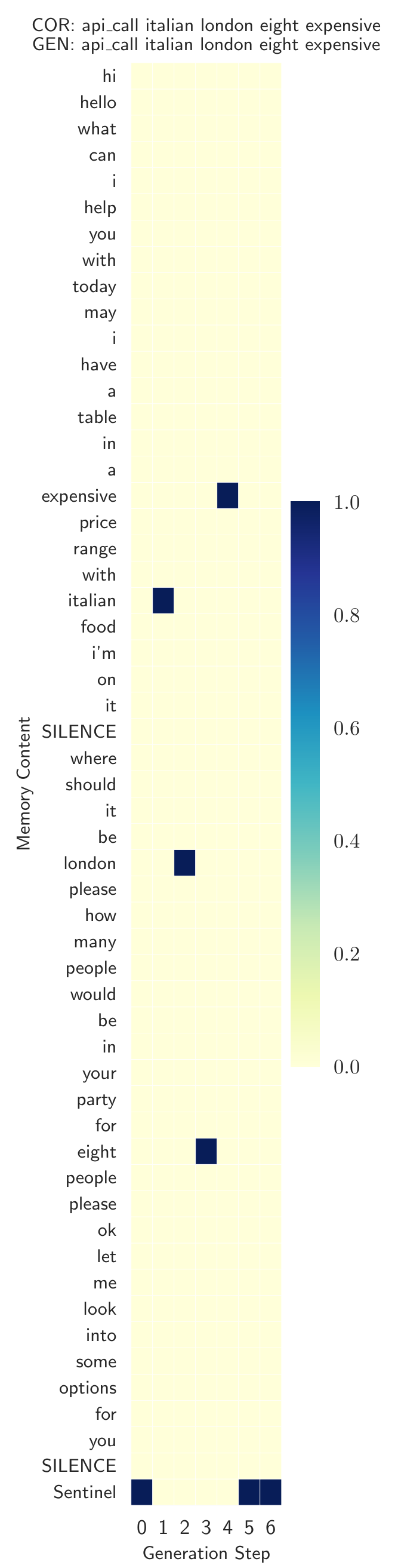}
\caption{Last hop memory attention visualization from the bAbI dataset. COR and GEN on the top are the correct response and our generated one.}
\label{MEMATT}
\end{figure}

\begin{figure}[H]
\centering
\includegraphics[width=\linewidth]{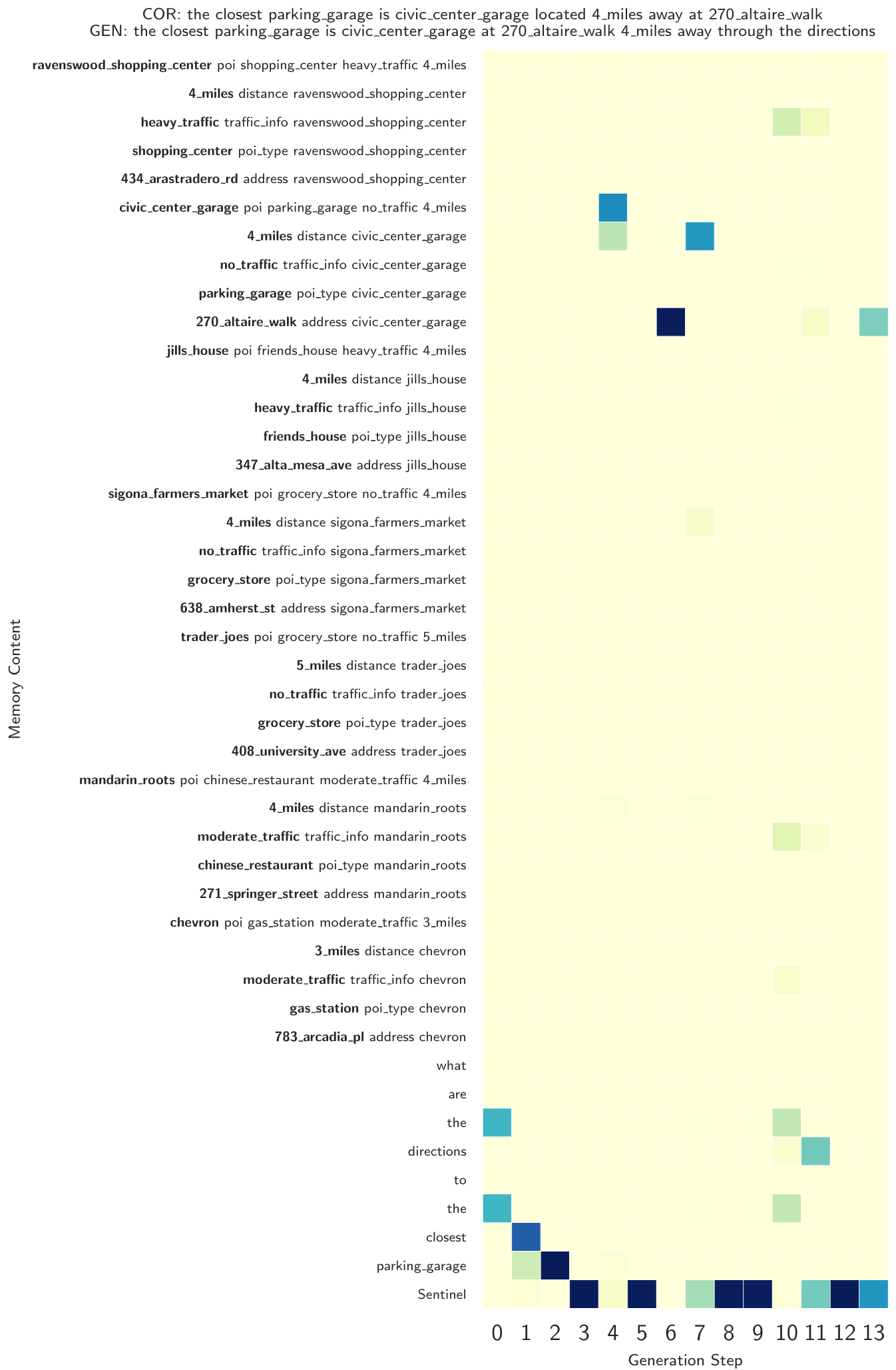}
\caption{Last hop memory attention visualization from the In-Car dataset. COR and GEN on the top are the correct response and our generated one.}
\label{MEMATT}
\end{figure}

\section{InCar Assistant Dataset Examples}


\begin{table}[H]
\centering
\caption{Example of generated responses for the In-Car Assistant on the weather domain.}
\label{APPENDINX_1}
\resizebox{\linewidth}{!}{
\begin{tabular}{|c|c|c|c|}
\hline
\textbf{location}                & \textbf{monday}                      & \textbf{tuesday}                  & \textbf{wednesday}                    \\ \hline
grand rapids                     & hot, low of 50F, high of 70F         & raining, low of 60F, high of 80F  & rain, low of 20F, high of 30F         \\ \hline
new york                         & misty, low of 30F, high of 50F       & snow, low of 70F, high of 80F     & cloudy, low of 20F, high of 30F       \\ \hline
boston                           & hail, low of 90F, high of 100F       & overcast, low of 60F, high of 70F & rain, low of 50F, high of 60F         \\ \hline
durham                           & hot, low of 90F, high of 100F        & dry, low of 60F, high of 80F      & misty, low of 60F, high of 80F        \\ \hline
san francisco                    & rain, low of 60F, high of 70F        & cloudy, low of 30F, high of 40F   & overcast, low of 90F, high of 100F    \\ \hline
carson                           & raining, low of 70F, high of 80F     & humid, low of 90F, high of 100F   & frost, low of 40F, high of 60F        \\ \hline
san jose                         & blizzard, low of 40F, high of 50F    & snow, low of 90F, high of 100F    & overcast, low of 60F, high of 80F     \\ \hline
\textbf{thursday}                    & \textbf{friday}                  & \textbf{saturday}                     & \textbf{sunday}                   \\ \hline
clear skies, low of 60F, high of 70F & warm, low of 70F, high of 90F    & foggy, low of 50F, high of 60F        & overcast, low of 50F, high of 60F \\ \hline
rain, low of 80F, high of 100F       & rain, low of 40F, high of 60F    & cloudy, low of 30F, high of 50F       & snow, low of 20F, high of 40F     \\ \hline
dew, low of 20F, high of 30F         & cloudy, low of 90F, high of 100F & overcast, low of 50F, high of 70F     & overcast, low of 80F, high of 90F \\ \hline
misty, low of 90F, high of 100F      & hot, low of 70F, high of 90F     & hail, low of 30F, high of 40F         & rain, low of 60F, high of 80F     \\ \hline
stormy, low of 90F, high of 100F     & cloudy, low of 90F, high of 100F & clear skies, low of 90F, high of 100F & foggy, low of 30F, high of 50F    \\ \hline
dry, low of 50F, high of 60F         & foggy, low of 80F, high of 100F  & overcast, low of 50F, high of 70F     & overcast, low of 60F, high of 70F \\ \hline
rain, low of 30F, high of 50F        & rain, low of 80F, high of 100F   & dry, low of 50F, high of 60F          & dry, low of 90F, high of 100F     \\ \hline
\end{tabular}
}

\begin{tabular}{r|l}
\hline
\textbf{DRIVER}      & Is it gonna snow in Durham during the week?                \\ \hline
\textit{Seq2seq}     & there is no snow in the forecast for this week             \\
\textit{+Att}        & there is no snow in the forecast for durham for durham     \\
\textit{Ptr-Unk}      & yes there is no snow in durham on week                     \\ \hline
\textit{Mem2Seq} & no snow, the forecast does not predict any snow on Durham  \\ \hline
\textbf{GOLD}        & Durham does not have any snow predicted this week          \\ \hline
\end{tabular}
\end{table}

\begin{table}[H]
\centering
\caption{Example of generated answers for the In-Car dataset on the Weather Domain.}
\label{APPENDINX_2}
\resizebox{\linewidth}{!}{
\begin{tabular}{|c|c|c|c|}
\hline
\textbf{monday}                      & \textbf{tuesday}                      & \textbf{friday}                      & \textbf{wednesday}                   \\ \hline
windy, low of 40F, high of 50F       & snow, low of 40F, high of 50F         & frost, low of 30F, high of 40F       & hail, low of 70F, high of 80F        \\ \hline
rain, low of 30F, high of 40F        & foggy, low of 60F, high of 70F        & stormy, low of 50F, high of 60F      & snow, low of 80F, high of 90F        \\ \hline
cloudy, low of 20F, high of 40F      & clear skies, low of 70F, high of 90F  & clear skies, low of 60F, high of 70F & dry, low of 40F, high of 60F         \\ \hline
warm, low of 90F, high of 100F       & windy, low of 50F, high of 60F        & blizzard, low of 20F, high of 30F    & rain, low of 60F, high of 70F        \\ \hline
snow, low of 60F, high of 80F        & clear skies, low of 30F, high of 50F  & clear skies, low of 30F, high of 40F & dry, low of 30F, high of 50F         \\ \hline
windy, low of 50F, high of 60F       & drizzle, low of 80F, high of 100F     & windy, low of 80F, high of 100F      & clear skies, low of 50F, high of 70F \\ \hline
clear skies, low of 60F, high of 70F & rain, low of 90F, high of 100F        & foggy, low of 20F, high of 40F       & hail, low of 20F, high of 30F        \\ \hline
\textbf{thursday}                    & \textbf{sunday}                       & \textbf{location}                    & \textbf{saturday}                    \\ \hline
raining, low of 80F, high of 90F     & foggy, low of 40F, high of 50F        & san francisco                        & snow, low of 40F, high of 50F        \\ \hline
drizzle, low of 50F, high of 60F     & clear skies, low of 90F, high of 100F & manhattan                            & dry, low of 70F, high of 90F         \\ \hline
stormy, low of 40F, high of 50F      & hail, low of 40F, high of 60F         & seattle                              & windy, low of 60F, high of 70F       \\ \hline
overcast, low of 80F, high of 90F    & blizzard, low of 20F, high of 40F     & oakland                              & clear skies, low of 50F, high of 70F \\ \hline
raining, low of 50F, high of 60F     & windy, low of 30F, high of 50F        & exeter                               & overcast, low of 60F, high of 70F    \\ \hline
frost, low of 40F, high of 50F       & drizzle, low of 40F, high of 60F      & menlo park                           & foggy, low of 50F, high of 70F       \\ \hline
dry, low of 20F, high of 30F         & cloudy, low of 20F, high of 40F       & carson                               & windy, low of 80F, high of 100F      \\ \hline
\end{tabular}
}

\begin{tabular}{r|l}
\hline
\textbf{DRIVER}      & Is it going to snow today?                         \\
\textbf{CAR}         & What city do you wanna know if it will snow today? \\
\textbf{DRIVER}      & Please check for Oakland                           \\
\textbf{GOLD}        & Today is gonna be warm in Oakland                  \\ \hline
\textit{Seq2seq}     & It is not snow in today in                         \\
\textit{+att}        & No there will be snow in Oakland today             \\
\textit{Ptr-Unk}      & Today is gonna be PAD in                           \\ \hline
\textit{Mem2Seq} & it will not snow in oakland today                  \\ \hline
\end{tabular}
\end{table}

\begin{table}[H]
\centering
\caption{Example of generated responses for the In-Car Assistant on the navigation domain.}
\label{APPENDINX_4}
\resizebox{\linewidth}{!}{
\begin{tabular}{|c|c|c|c|c|}
\hline
\textbf{poi} & \textbf{distance} & \textbf{traffic\_info} & \textbf{poi\_type} & \textbf{address} \\ \hline
Stanford Express Care & 3 miles & no traffic & hospital & 214 El Camino Real \\ \hline
home & 4 miles & heavy traffic & home & 5671 barringer street \\ \hline
Whole Foods & 2 miles & heavy traffic & grocery store & 819 Alma St \\ \hline
P.F. Changs & 6 miles & moderate traffic & chinese restaurant & 669 El Camino Real \\ \hline
Jing Jing & 5 miles & heavy traffic & chinese restaurant & 113 Arbol Dr \\ \hline
Pizza My Heart & 6 miles & moderate traffic & pizza restaurant & 528 Anton Ct \\ \hline
Cafe Venetia & 4 miles & no traffic & coffee or tea place & 269 Alger Dr \\ \hline
\end{tabular}
}

\begin{tabular}{r|l}
\hline
\textbf{DRIVER} & Find me a tea house \\\hline
\textit{Seq2seq} & We' re 2 miles away from from and there is a road block nearby \\
\textit{+att} & We' re 4 miles away from Cafe Venetia and it s at 269 Alger drive \\
\textit{PTRUNK} & We' re 6 miles away from whole foods and you like the address on screen \\ \hline
\textit{Mem2Seq} & We're 4 miles away from Cafe Venetia and there is no traffic \\ \hline
\textbf{GOLD} & There is Cafe Venetia that's 4 miles away at 269 Alger dr \\ \hline
\end{tabular}
\end{table}

\begin{table}[H]
\centering
\caption{Example of generated responses for the In-Car Assistant on the navigation domain.}
\label{APPENDINX_5}
\resizebox{\linewidth}{!}{
\begin{tabular}{|c|c|c|c|c|}
\hline
\textbf{distance} & \textbf{traffic\_info} & \textbf{poi\_type} & \textbf{address} & \textbf{poi} \\ \hline
4 miles & heavy traffic & rest stop & 578 Arbol Dr & Hotel Keen \\ \hline
1 miles & moderate traffic & hospital & 899 Ames Ct & Stanford Childrens Health \\ \hline
2 miles & no traffic & grocery store & 452 Arcadia Pl & Safeway \\ \hline
1 miles & moderate traffic & coffee or tea place & 436 Alger Dr & Palo Alto Cafe \\ \hline
1 miles & heavy traffic & grocery store & 819 Alma St & Whole Foods \\ \hline
6 miles & moderate traffic & coffee or tea place & 394 Van Ness Ave & Coupa \\ \hline
6 miles & heavy traffic & coffee or tea place & 145 Amherst St & Teavana \\ \hline
\end{tabular}
}
\resizebox{\linewidth}{!}{
\begin{tabular}{r|l}
\hline
\textbf{DRIVER}      & Take me to a place where i can get some hot tea                                               \\\hline
\textit{Seq2seq}     & There is a away                                                                               \\
\textit{att}         & There is a Pizza Chicago and Teavana                                                          \\
\textit{PTRUNK}      & I have a PAD away away                                                                        \\ \hline
\textit{Mem2Seq} & We' re 6 miles away from Palo Alto Cafe and 6 miles away from Coupa, which one do you prefer? \\ \hline
\textbf{GOLD}        & Coupa is nearby and serves tea  \\ \hline
\end{tabular}}
\end{table}

\begin{table}[H]
\centering
\caption{Example of generated responses for the In-Car Assistant on the navigation domain.}
\label{APPENDINX_6}
\resizebox{\linewidth}{!}{
\begin{tabular}{|c|c|c|c|c|}
\hline
\textbf{distance} & \textbf{traffic\_info} & \textbf{poi\_type} & \textbf{address}   & \textbf{poi}   \\ \hline
3 miles           & no traffic             & chinese restaurant & 593 Arrowhead Way  & Chef Chu's     \\ \hline
1 miles           & no traffic             & chinese restaurant & 669 El Camino Real & P.F. Changs    \\ \hline
5 miles           & road block nearby      & gas station        & 200 Alester Ave    & Valero         \\ \hline
3 miles           & no traffic             & pizza restaurant   & 528 Anton Ct       & Pizza My Heart \\ \hline
4 miles           & heavy traffic          & rest stop          & 753 University Ave & Comfort Inn    \\ \hline
5 miles           & heavy traffic          & chinese restaurant & 842 Arrowhead Way  & Panda Express  \\ \hline
2 miles           & heavy traffic          & pizza restaurant   & 704 El Camino Real & Pizza Hut      \\ \hline
6 miles           & no traffic             & friends house      & 864 Almanor Ln     & jacks house    \\ \hline
\end{tabular}
}
\resizebox{\linewidth}{!}{
\begin{tabular}{r|l}
\hline
\textbf{DRIVER} & Find me the nearest gas station. \\\hline
\textit{Seq2seq} & the nearest parking garage is from at away \\
\textit{+att} & The is one is away away \\
\textit{PTRUNK} & PAD nearest gas station is is is \\ \hline
\textit{Mem2Seq} & The nearest gas station is gas station which is you you \\ \hline
\textbf{GOLD} & The nearest gas station is located 5 miles away. Need more info? \\ \hline
\end{tabular}
}
\end{table}

\begin{table}[H]
\centering
\caption{Example of generated responses for the In-Car Assistant on the navigation domain.}
\label{APPENDINX_9}
\resizebox{\linewidth}{!}{
\begin{tabular}{|c|c|c|c|c|}
\hline
\textbf{Poi}             & \textbf{Distance} & \textbf{Traffic\_info} & \textbf{Poi\_type} & \multicolumn{1}{c|}{\textbf{Address}} \\ \hline
The Westin               & 4 miles           & no traffic             & rest stop          & 329 El Camino Real                    \\ \hline
Palo Alto Garage R       & 2 miles           & moderate traffic       & parking garage     & 481 Amaranta Ave                      \\ \hline
Mandarin Roots           & 4 miles           & moderate traffic       & chinese restaurant & 271 Springer Street                   \\ \hline
Sigona Farmers Market    & 4 miles           & no traffic             & grocery store      & 638 Amherst St                        \\ \hline
Whole Foods              & 2 miles           & heavy traffic          & grocery store      & 819 Alma St                           \\ \hline
Stanford Shopping Center & 2 miles           & moderate traffic       & shopping center    & 773 Alger Dr                          \\ \hline
toms house               & 1 miles           & no traffic             & friends house      & 580 Van Ness Ave                      \\ \hline
\end{tabular}}
\resizebox{\linewidth}{!}{
\begin{tabular}{r|l}
\hline
\textbf{DRIVER}        & Give me directions to the closest grocery store                                                                                               \\
\textbf{CAR}           & \begin{tabular}[c]{@{}l@{}}There are Whole Foods 2 miles away and \\ Sigona Farmers Market 4 miles away. Where do we go?\end{tabular}         \\
\textbf{DRIVER}        & I need the one that avoids all heavy traffic                                                                                                  \\ \hline
\textit{Seq2seq}       & he is is is is away at                                                                                                                       \\
\textit{+Att}           & Chef Chu s is located at 593 Arrowhead way                                                                                                    \\
\textit{Ptr-Unk}        & PAD is the is PAD the to                                                                                                                      \\ \hline
\textit{Mem2Seq}   & \begin{tabular}[c]{@{}l@{}}Sigona Farmers Market is the only option i is 4 miles \\ away with no traffic, setting navigation now\end{tabular} \\ \hline
\textbf{GOLD} & \begin{tabular}[c]{@{}l@{}}Sigona Farmers Market is located 4 miles away with no traffic. \\ It's located at 638 Amherst St.\end{tabular}     \\ \hline
\end{tabular}}
\end{table}

\begin{table}[H]
\centering
\caption{Example of generated responses for the In-Car Assistant on the scheduling domain.}
\label{APPENDINX_7}
\resizebox{\linewidth}{!}{
\begin{tabular}{r|l}
\hline
\textbf{DRIVER} & Set me a reminder for football activity at 1pm this month on the 11th with marie \\\hline
\textit{Seq2seq} & Okay I am scheduling a reminder with your for at on with at \\
\textit{att} & Okay I am scheduling a football activity with on the 15th at 1pm this month on the 15th \\
\textit{PTRUNK} & Okay setting PAD reminder reminder football activity football activity for this this on the 11th month 1pm \\ \hline
\textit{Mem2Seq} & Setting a reminder for football activity with marie on the 11th at 1pm \\ \hline
\textbf{GOLD} & Reminder for football on the 11th of this month at 1pm with marie is set \\ \hline
\end{tabular}
}

\end{table}

\begin{table}[H]
\centering
\caption{Example of generated responses for the In-Car Assistant on the scheduling domain.}
\label{APPENDINX_8}
\resizebox{\linewidth}{!}{
\begin{tabular}{|c|c|c|c|c|c|}
\hline
\textbf{room} & \textbf{agenda} & \textbf{time} & \textbf{date} & \textbf{party} & \textbf{event} \\ \hline
conference room 50 & discuss the merger & 1pm & Wednesday & infrastructure team & conference \\ \hline
- & - & 2pm & Wednesday & - & yoga activity \\ \hline
- & - & 2pm & Wednesday & - & doctor appointment \\ \hline
conference room 50 & go over budget & 7pm & Monday & infrastructure team & meeting \\ \hline
- & - & 5pm & Monday & - & dinner \\ \hline
- & - & 10am & Thursday & - & medicine \\ \hline
\end{tabular}
}
\resizebox{\linewidth}{!}{
\begin{tabular}{r|l}
\hline
\textbf{DRIVER} & Tell me when is my next meeting and who are invited in the meeting and what are the agenda of the next meeting \\\hline
\textit{Seq2seq} & The meeting is the meeting in conference room 100 at 1pm \\
\textit{+att} & Your next meeting is on wednesday at 1pm \\
\textit{PTRUNK} & your PAD meeting PAD the in in the to to the the the the \\ \hline
\textit{Mem2Seq} & Your next meeting is on Monday at 7pm hr is to hr in attendance to go over the budget \\ \hline
\textbf{GOLD} & Your next meeting is on Monday at 7 pm with the infrastructure team to go over budget \\ \hline
\end{tabular}
}
\end{table}

\end{document}